# Learning in Real-Time Search: A Unifying Framework

**Vadim Bulitko**                                              BULITKO@UALBERTA.CA
**Greg Lee**                                                   GREGLEE@CS.UALBERTA.CA
*Department of Computing Science*
*University of Alberta*
*Edmonton, Alberta T6G 2E8, CANADA*

## Abstract

Real-time search methods are suited for tasks in which the agent is interacting with an initially unknown environment in real time. In such simultaneous planning and learning problems, the agent has to select its actions in a limited amount of time, while sensing only a local part of the environment centered at the agent's current location. Real-time heuristic search agents select actions using a limited lookahead search and evaluating the frontier states with a heuristic function. Over repeated experiences, they refine heuristic values of states to avoid infinite loops and to converge to better solutions. The wide spread of such settings in autonomous software and hardware agents has led to an explosion of real-time search algorithms over the last two decades. Not only is a potential user confronted with a hodgepodge of algorithms, but he also faces the choice of control parameters they use. In this paper we address both problems. The first contribution is an introduction of a simple three-parameter framework (named LRTS) which extracts the core ideas behind many existing algorithms. We then prove that LRTA*, $\epsilon$-LRTA* , SLA*, and $\gamma$-Trap algorithms are special cases of our framework. Thus, they are unified and extended with additional features. Second, we prove completeness and convergence of any algorithm covered by the LRTS framework. Third, we prove several upper-bounds relating the control parameters and solution quality. Finally, we analyze the influence of the three control parameters empirically in the realistic scalable domains of real-time navigation on initially unknown maps from a commercial role-playing game as well as routing in *ad hoc* sensor networks.

## 1. Motivation

In this paper, we consider a simultaneous planning and learning problem. One motivating application lies with navigation on an initially unknown map under real-time constraints. As an example, consider a robot driving to work every morning. Imagine the robot to be a newcomer to the town. The first route the robot finds may not be optimal because traffic jams, road conditions, and other factors are initially unknown. With the passage of time, the robot continues to learn and eventually converges to a nearly optimal commute. Note that planning and learning happen while the robot is driving and therefore are subject to time constraints.

Present-day mobile robots are often plagued by localization problems and power limitations, but their simulation counter-parts already allow researchers to focus on the planning and learning problem. For instance, the RoboCup Rescue simulation league (Kitano, Tadokoro, Noda, Matsubara, Takahashi, Shinjou, & Shimada, 1999) requires real-time planning and learning with multiple agents mapping out an unknown terrain. Pathfinding is done in real time as various crises, involving fire spread and human victims trapped in rubble, progress while the agents plan.

Similarly, many current-generation real-time strategy games employ *a priori* known maps. Full knowledge of the maps enables complete search methods such as A* (Hart, Nilsson, & Raphael, 1968) and Dijkstra's algorithm (Dijkstra, 1959). Prior availability of the maps allows pathfinding engines to pre-compute various data to speed up on-line navigation. Examples of such data include visibility graphs (Woodcock, 2000), influence maps (Pottinger, 2000), space triangulation (Kallmann, Bieri, & Thalmann, 2003), state abstraction hierarchies (Holte, Drummond, Perez, Zimmer, & MacDonald, 1994; Holte, 1996; Botea, Müller, & Schaeffer, 2004) and route waypoints (Reece, Krauss, & Dumanoir, 2000). However, the forthcoming generations of commercial and academic games (Buro, 2002) will require the agent to cope with initially unknown maps via exploration and learning *during* the game, and therefore will greatly limit the applicability of complete search algorithms and pre-computation techniques.

Incremental search methods such as dynamic A* (D*) (Stenz, 1995) and D* Lite (Koenig & Likhachev, 2002) can deal with initially unknown maps and are widely used in robotics, including DARPA's Unmanned Ground Vehicle program, Mars rover, and other mobile robot prototypes (Herbert, McLachlan, & Chang, 1999; Thayer, Digney, Diaz, Stentz, Nabbe, & Hebert, 2000). They work well when the robot's movements are slow with respect to its planning speed (Koenig, 2004). In real-time strategy games, however, the AI engine can be responsible for hundreds to thousands of agents traversing a map simultaneously and the planning cost becomes a major factor. To illustrate: even at the smaller scale of the six-year old "Age of Empires 2" (Ensemble-Studios, 1999), 60-70% of simulation time is spent in pathfinding (Pottinger, 2000). This gives rise to the following questions:

1. How can planning time per move, and particularly the first-move delay, be minimized so that each agent moves smoothly and responds to user requests nearly instantly?

2. Given real-time execution, local sensory information, and initially unknown terrain, how can the agent learn a near-optimal path and, at the same time, minimize the learning time and memory required?

The rest of the paper is organized as follows. We first introduce a family of real-time search algorithms designed to address these questions. We then make the first contribution by defining a simple parameterized framework that unifies and extends several popular real-time search algorithms. The second contribution lies with a theoretical analysis of the resulting framework wherein we prove convergence and completeness as well as several performance bounds. Finally, we evaluate the influence of the control parameters in two different domains: single agent navigation in unknown maps and routing in *ad hoc* sensor networks. Detailed pseudocode needed to reimplement the algorithms as well as to follow the theorem proofs is presented in Appendix A. Theorem proofs themselves are found in Appendix B.

## 2. Real-time Heuristic Search and Related Work

We situate our survey of real-time heuristic search literature in the framework of *agent-centered search* (Koenig, 2001). While traditional off-line search methods first plan a path from start to goal state and then move the agent along the path, agent-centered search interleaves planning and execution. Furthermore, the planning is restricted to the areas of the search space around the current state of the agent such as the physical location of a mobile robot or the current board position in a game. The agent-centered search methods thus satisfy these two requirements: (i) at all times, the agent is in a single state which can be changed only via taking actions and, therefore, incurring execution cost; (ii) the agent can see only the states around its current state.

*Real-time heuristic search* methods are a subset of agent-centered search methods. They are distinguished by the following two additional properties: (i) the planning time per move can be upper-bounded by a user-supplied constant (hence *real-time*); (ii) they associate a heuristic function with each state (hence *heuristic*). The former requirement is motivated by the need to act quickly in control tasks such as flying a helicopter (Ng, Coates, Diel, Ganapathi, Schulte, Tse, Berger, & Liang, 2004) or playing a real-time strategy computer game (Buro, 2002). The latter property allows the agent to avoid infinite cycles on a map by learning the proper distances from each state to the goal state. Unlike learning the map and using it for planning (Stenz, 1995; Koenig & Likhachev, 2002; Chimura & Tokoro, 1994; Sturtevant, 2005), acquiring high-quality heuristic values lets the agent select the right action very quickly thereby improving its reaction time. Having the heuristic values also allows the agent to pick the next best action should the primary choice become unavailable (e.g., due to a road block on the map). Remarkably, learning state and state-action values is the prevailing technique in Reinforcement Learning (Sutton, 2005).

Learning real-time search algorithms, such as LRTA* (Korf, 1990), interleave planning and execution in an on-line decision-making setting. As the planning time per each action executed by the agent is bounded, these algorithms can be used as control policies for autonomous agents, even in an unknown and/or non-stationary environment (Koenig & Simmons, 1998; Koenig, 1999; Ishida & Korf, 1991, 1995; Ishida, 1997). In particular, the ability to make decisions after only a small-scale local search makes these algorithms well suited for routing in large-scale wireless networks of simple sensors and actuators where each node is aware only of its nearby neighbors and no global map exists. In such scenarios, not only does the limited amount of computation per node suit the low computing power and energy requirements, but also the network collectively learns its own topology over time (Yu, Govindan, & Estrin, 2001; Shang, Fromherz, Zhang, & Crawford, 2003).

Since the pioneering LRTA* (Korf, 1990), research in the field of learning real-time heuristic search has developed in several major directions. Ishida and Korf (1991) investigated modifications to LRTA* for non-stationary environments. Shimbo and Ishida (2003) studied convergence to suboptimal solutions as well as mechanisms for bounding the amount of state space exploration. Furcy and Koenig (2000) considered a different learning mechanism speeding up the convergence. Russell and Wefald (1991) researched a decision-theoretic approach to balancing partial planning and execution. Shue and Zamani (1993) and Bulitko (2004) proposed a backtracking component suggesting yet another way to control the exploration of the environment.

Note that while the original LRTA* algorithm can be viewed as a special case of real-time dynamic programming (Barto, Bradtke, & Singh, 1995), in general, real-time heuristic search methods have several notable differences from reinforcement learning methods. First, they usually assume a deterministic environment and thus take advantage of more aggressive value update rules. Second, they employ non-trivial initial heuristics and can converge even faster when such a heuristic is admissible (Pearl, 1984) by never decreasing heuristic values of states. Third, real-time heuristic search methods often use more extensive and sophisticated local search methods than the $\epsilon$-greedy control policy commonly used in reinforcement learning. Examples include dynamically selected lookahead search space in DTA* (Russell & Wefald, 1991), additional heuristic for tie-breaking in FALCONS (Furcy & Koenig, 2000), and heuristic upper-bounds for safe space exploration in bounded LRTA* (Shimbo & Ishida, 2003). This improves the quality of decision-making by compensating for inaccuracies in the heuristic function and speeds up the learning process. Finally, backtracking extensions (Shue & Zamani, 1993; Bulitko, 2004) give the agent another mechanism to maintain consistency of its heuristic.

The multitude of learning real-time search algorithms (LRTA*, $\epsilon$-LRTA* , $\delta$-LRTA*, FAL-CONS, eFALCONS, SLA*, $\gamma$-Trap, SLA*T, DTA*, etc.) available to the user can be disorienting since he has to decide not only on the algorithm to use but also on the algorithm parameters which can have a major impact on performance. The problem is further complicated by the fact that the empirical studies have been done in different test beds making the results incomparable directly. To illustrate: Furcy and Koenig (2000) and Shimbo and Ishida (2003) appear to use the same testbed (the 8-puzzle). Yet, their results on the same algorithm (LRTA*) differ substantially. A closer inspection reveals that two different goal states were used leading to the major discrepancies in the performance. To compound these problems, most performance metrics have been measured with the myopic lookahead search of depth one. This is in contrast to game-playing practice where most competitive systems gain a substantial benefit from a deeper search horizon (Schaeffer, Culberson, Treloar, Knight, Lu, & Szafron, 1992; Hsu, Campbell, & Hoane, 1995; Buro, 1995).

We take a step towards a unified view of learning in real-time search and make four contributions. First, we introduce a simple three-parameter framework (named LRTS) that includes LRTA*, $\epsilon$-LRTA* , SLA* and $\gamma$-Trap as its special cases. A by-product of this generalization is an extension of the first three algorithms to variable depth lookahead. Second, we prove completeness and convergence for any combination of the parameters. Third, we prove non-trivial theoretical bounds on the influence of the parameters. Fourth, we present a large-scale empirical evaluation in practically relevant domains.

## 3. Problem Formulation

In this section, we formally introduce the learning real-time search problem settings and the metrics we will be using.

**Definition 3.1** The search space is defined as a tuple $(S, A, c, s_0, S_g, h_0)$ where $S$ is a finite set of states, $A$ is a finite set of actions, $c : S \times A \to \mathbb{R}^+$ is the cost function with $c(s, a)$ being the incurred cost of taking action $a$ in state $s$, $s_0$ is the initial state, $S_g \subset S$ is the set of goal states, and $h_0$ is the initial heuristic (e.g., Manhattan distance).

We adopt the assumptions of Shue and Zamani (1993), Shimbo and Ishida (2003) that (i) for every action in every state there exists a reverse action (possibly with a different cost), (ii) every applicable action leads to another state (i.e., no self-loops), and (iii) at least one goal state in $S_g$ is reachable from $s_0$. Assumption (i) is needed for the backtracking mechanism used in SLA* (Shue & Zamani, 1993). It holds in many combinatorial and path-planning problems. Its applicability to domains where reversing an action may require a non-trivial sequence of actions is a subject of future research. Assumption (ii) is adopted for the sake of proof simplicity. In other words, the results presented in this paper hold even if self-loops (of positive cost) are present. As with all real-time search algorithms we are unifying, we assume that the environment is stationary and deterministic. Extension of our algorithms on dynamic/stochastic domains is a subject of current research.

**Definition 3.2** The execution cost of traveling from state $s_1$ to state $s_2$ denoted by $\mathrm{dist}(s_1, s_2)$ is defined as the minimal cumulative execution cost the agent is to incur by traveling from $s_1$ to $s_2$. Throughout the paper, we will assume that $\mathrm{dist}$ satisfies the standard triangle inequality: $\forall s_1, s_2, s_3 \in S \left[ \mathrm{dist}(s_1, s_3) \le \mathrm{dist}(s_1, s_2) + \mathrm{dist}(s_2, s_3) \right]$. Then, for any state $s$ its true execution cost $h^*$ is defined as the minimal execution cost to the nearest goal: $h^*(s) = \min_{s_g \in S_g} \text{dist}(s, s_g)$. A heuristic approximation $h$ to the true execution cost is called admissible iff $\forall s \in S \, [h(s) \le h^*(s)]$. The value of $h$ in state $s$ will be referred to as the heuristic value of state $s$. We will assume that the heuristic value of the initial heuristic function is 0 for any goal state. The latter assumption holds trivially if the initial heuristic function is admissible. A depth $d$ child of state $s$ is any state $s'$ reachable from $s$ in the minimum of $d$ actions (denoted by $\|s, s'\| = d$). The depth $d$ neighborhood of state $s$ is then defined as $S(s, d) = \{s_d \in S \mid \|s, s_d\| = d\}$.

**Definition 3.3** Search agents operate by starting in a fixed start state $s_0$ and executing actions suggested by their control policies. A *trial* is defined as a sequence of states the algorithm visits between the start state and the first goal state it encounters. Once a trial is completed, the agent is reset to the start state and the next trial begins. *Final trial* is defined as the first trial on which no learning (i.e., updates to the heuristic function) occurs.[1] A *convergence run* is the sequence of trials from the first trial to the final trial.

## 4. Performance Measures

Each problem instance is fully specified by the search space which includes the start and goal states. The agent is run until convergence and the following statistics are collected:

**execution convergence cost** is the sum of execution costs of the actions taken by the agent during the convergence process (i.e., on the convergence run);

**planning cost** is the average cost of planning per action during the convergence process. Planning cost of an action is the number of states the agent considered to decide on taking the action;

**total convergence cost** is the total planning convergence cost for all actions during convergence plus the execution convergence cost scaled by a factor called "planning speed". The scaling factor represents the amount of planning (measured in the number of nodes considered) the agent would be able to do in the time it takes to execute a unit of travel. For instance, the computer on-board an AIBO robodog may be able to plan with 10,000 states in the time it takes the AIBO to traverse 1 foot of distance on the ground. Correspondingly, the planning speed will be 10,000 states per foot. This is a commonly used way of combining the execution and planning costs into a single metric (Russell & Wefald, 1991; Koenig, 2004);[2]

**memory** is the total amount of memory (measured in the number of state values) the agent used to store the heuristic function during the convergence run. Note that the initial heuristic is usually represented in a compact algorithmic form as opposed to a table. Therefore, memory is required to store only the heuristic values modified during learning;

**first-move delay (lag)** is the amount of planning time (measured in milliseconds[3]) the agent takes before deciding on the first move. This is an important metric in real-time strategy games,

---

1. Note that if random tie-breaking is used, more learning can actually take place after a learning-free trial. We use fixed tie-breaking throughout this paper for simplicity.

2. Note that unlike total planning cost, total convergence cost allows us to model domains where some actions are expensive disproportionally to their running time (e.g., taking damage by running into a wall). Additionally, we use the standard real-time search framework and assume that planning and execution are not simultaneous but interleaved.

3. All timings are taken on a PowerMac G5 running at 2GHz. Apple gcc 3.3 compiler is used.

wherein hundreds to thousands of units can be tasked simultaneously and yet have to react to user's commands as quickly as possible;

**suboptimality of the final solution** is defined as the percentage by which the execution cost of the *final* trial exceeds that of the best possible solution. For instance, if a pathfinding agent had the execution cost of 120 on the final trial and the shortest path has the cost of 100, the suboptimality is 20%.

## 5. Application Domains

In the paper, we illustrate the algorithms in two realistic testbeds: real-time navigation in unknown terrain and routing in *ad hoc* wireless sensor networks. The former domain is more standard to single-agent search and so we will use it throughout the paper to illustrate our points. The latter domain is a relative newcomer to the field of real-time heuristic search. We will use it later in the paper (Section 8) to demonstrate how learning single-search methods can handle a practically important yet different task.

Note that both domains have the desired attributes listed in the introduction. Indeed, in both domains, the state space is initially unknown, the agent is in a single current state to be changed via executing actions, the agent can sense only a local part of the state space centered on the agent's current state, the planning time per action should be minimized, and there are repeated trials.

### 5.1 Real-time navigation in next-generation video games

The agent is tasked to travel from the start state $(x_s, y_s)$ to the single goal state $(x_g, y_g)$. The coordinates are on a two-dimensional rectangular grid. In each state, up to eight moves are available leading to the eight immediate neighbors. Each straight move (i.e., north, south, west, east) has a *travel cost* of 1 while each diagonal move has a travel cost of $\sqrt{2}$. Each state on the map can be passable or occupied by a wall. In the latter case, the agent is unable to move into it. Initially, the map in its entirety is unknown to the agent. In each state $(x, y)$ the agent can see the status (occupied/free) in the neighborhood of the *visibility radius* $v$: $\{(x', y') \mid |x' - x| \leq v \ \& \ |y' - y| \leq v\}$. The initial heuristic we use is the so-called *octile distance* defined as the length of a shortest path between the two locations if all cells are passable. It is a translation of Manhattan distance onto the case of eight moves and can be easily computed in a closed form.

Note that classical A* search is inapplicable due to an initially unknown map. Specifically, it is impossible for the agent to plan its path through state $(x, y)$ unless it is either positioned within the visibility radius of the state or has visited this state on a prior trial. A simple solution to this problem is to generate the initial path under the assumption that the unknown areas of the map contain no occupied states (the free space assumption of Koenig, Tovey, & Smirnov, 2003). With the octile distance as the heuristic, the initial path is close to the straight line since the map is assumed to be empty. The agent follows the existing path until it runs into an occupied state. During the travel, it updates the explored portion of the map in its memory. Once the current path is blocked, A* is invoked again to generate a new complete path from the current position to the goal. The process repeats until the agent arrives at the goal. It is then reset to the start state and a new trial begins. The convergence run ends when no new states are seen.[4]

---

4. Note that this framework can be easily extended to the case of multiple start and goal states.

To increase planning efficiency, several methods of re-using information over subsequent planning episodes have been suggested. Two popular versions are D* (Stenz, 1995) and D* Lite (Koenig & Likhachev, 2002). Powerful as they are, these enhancements do not reduce the first-move lag time. Specifically, after the agent is given the destination coordinates, it has to conduct an A* search from its position to the destination before it can move. Even on small maps, this delay can be substantial. This is in contrast to LRTA* (Korf, 1990), which only performs a small local search to select the first move. As a result, several orders of magnitude more agents can respond to a user's request in the time it takes one A* agent.

As present-day games circumvent these problem by using fully observable maps and pre-computing auxiliary data structures beforehand, we conduct our experiments in a research prototype, called Hierarchical Open Graph (HOG), developed at the University of Alberta by Nathan Sturtevant. It allows one to load maps from commercial role-playing and real-time strategy games such as Baldur's Gate (BioWare, 1999) and WarCraft III (Blizzard, 2002). We use five maps ranging in size from $214 \times 192$ states (of which 2765 are traversable) to $235 \times 204$ (with 16142 traversable states). The largest map is shown in Figure 1. A total of 1000 problem instances on the five maps were randomly chosen so that their shortest path lengths fall in the ten bins: 1-10, 11-20, ..., 91-100 with 100 problems per bin. We will use this suite of 1000 problems throughout the paper.

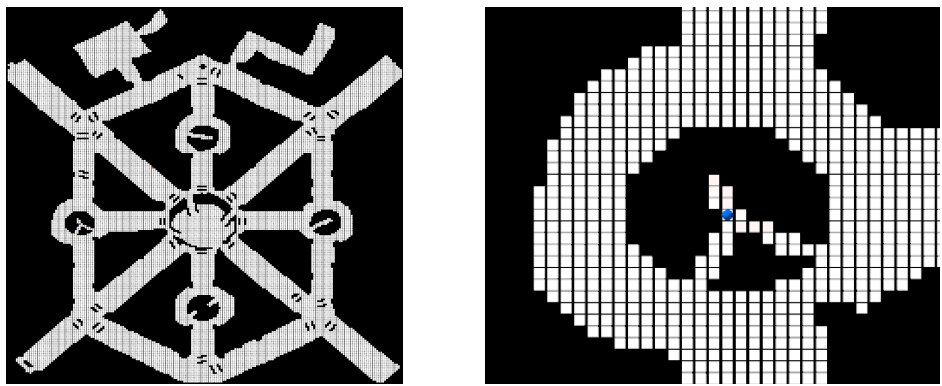

Figure 1: **Left:** one of the five test maps from Baldur's Gate – a commercial role-playing game by BioWare. **Right:** a close-up showing the agent as a dot in the middle.

## 6. LRTS: A Unifying Framework

In this section we will introduce the primary contribution of this paper — the unifying framework of real-time heuristic search — in an incremental fashion. Namely, we will start with the base algorithm, LRTA*, and then analyze three extensions: deeper lookahead, optimality weight, and backtracking. Each extension is illustrated with a hand-traced micro-example and empirical results in the real-time pathfinding domain. A unifying algorithm constitutes the section's finale.

### 6.1 Learning Real-Time A* (LRTA*)

LRTA* introduced by Korf (1990), is the first and best known learning real-time heuristic search algorithm . The key idea lies with interleaving acting and backing up heuristic values (Figure 2).[5]

_______________

5. For clarity, all pseudocode in this paper describes a single trial only.

Specifically, in the current state $s$, LRTA* with a lookahead of one considers the immediate neighbors (lines 4-5 in Figure 2). For each neighbor state, two values are computed: the execution cost of getting there from the current state (henceforth denoted by $g$) and the estimated execution cost of getting to the closest goal state from the neighbor state (henceforth denoted by $h$). While $g$ is known precisely, $h$ is a heuristic estimate. LRTA* then travels to the state with the lowest $f = g + h$ value (line 7). Additionally, it updates the heuristic value of the current state if the minimum $f$-value is greater than the heuristic value of the current state (line 6).[6]

---

**LRTA***

| | |
|---|---|
| 1 | initialize the heuristic: $h \leftarrow h_0$ |
| 2 | reset the current state: $s \leftarrow s_{\text{start}}$ |
| 3 | **while** $s \notin S_g$ **do** |
| 4 | generate children one move away from state $s$ |
| 5 | find the state $s'$ with the lowest $f = g + h$ |
| 6 | update $h(s)$ to $f(s')$ if $f(s')$ is greater |
| 7 | execute the action to get to $s'$ |
| 8 | **end while** |

---

Figure 2: LRTA* algorithm with a lookahead of one.

Korf (1990) showed that in finite domains where the goal state is reachable from any state and all action costs are non-zero, LRTA* finds a goal on every trial. Additionally, if the initial heuristic function is admissible then LRTA* will converge to an optimal solution after a finite number of trials. Compared to A* search, LRTA* with a lookahead of one has a considerably shorter first-move lag and finds suboptimal solutions much faster. However, converging to an optimal (e.g., lowest execution cost) solution can be expensive in terms of the number of trials and the total execution cost. Table 1 lists measurements averaged over 1000 convergence runs on the five game maps in the pathfinding domain (details in Section 5.1). The differences become more pronounced for larger problems (Figure 3).

Table 1: LRTA* with a lookahead of one vs. A* in pathfinding.

| Algorithm | First-move lag | Convergence execution cost |
|:---:|:---:|:---:|
| A* | 3.09 ms | 159 |
| LRTA* | 0.05 ms | 9346 |

## 6.2 Extension 1: Deeper Lookahead Search

LRTA* follows the heuristic landscape to a local minimum. It avoids getting stuck there by raising the heuristic values and eventually eliminating the local minimum. Thus, local heuristic minima, caused by inaccuracies in the initial heuristic values, are eliminated through the process of learning. Ishida (1997) referred to such inaccuracies in the heuristic as "heuristic depressions". He studied them for the basic case of LRTA* with a lookahead of one. Heuristic depressions were later generalized for the case of weighted LRTA* and arbitrary lookahead in (Bulitko, 2004) under the name of $\gamma$-traps.

---

6. This condition is not necessary if the heuristic is consistent as $f$-values will be non-decreasing along all lookahead branches. In general, there is no need to decrement the $h$-value of any state if the initial heuristic is admissible.

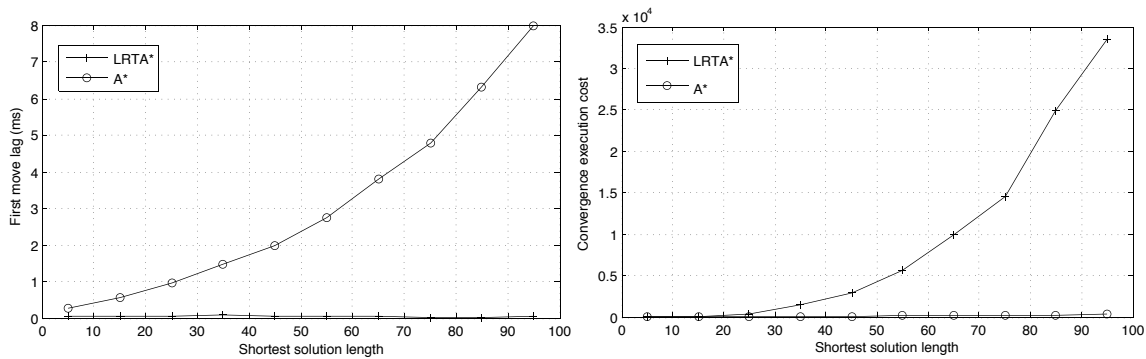

Figure 3: Differences between LRTA* and A* as pathfinding problems scale up.

In the process of "filling in" heuristic depressions, the agent can incur substantial execution cost by moving back and forth *inside* the depression. This happens due to the fact that LRTA* with a lookahead of one is myopic and conducts a minimum amount of planning per move. If planning is cheaper than execution, then a natural solution is to increase the amount of planning per move with the hope of eliminating heuristic local minima at a lower execution cost.

In LRTA*, additional planning per move is implemented through deeper lookahead search. The heuristic update and action selection rules introduced in the previous section can be extended to an arbitrary lookahead depth in a manner inspired by the standard mini-max search in games (Korf, 1990). Korf called the new rule "mini-min" and empirically observed that deeper lookahead decreases the execution cost but increases the planning cost per move. The phenomenon can be illustrated with a hand-traceable example in Figure 4. Each of the six states is shown as a circle with the actions shown as edges. Each action has the cost of 1. The initial heuristic $h_0$ is admissible but inaccurate. Heuristic values before and after the first trial are shown as numbers under each state (circle).

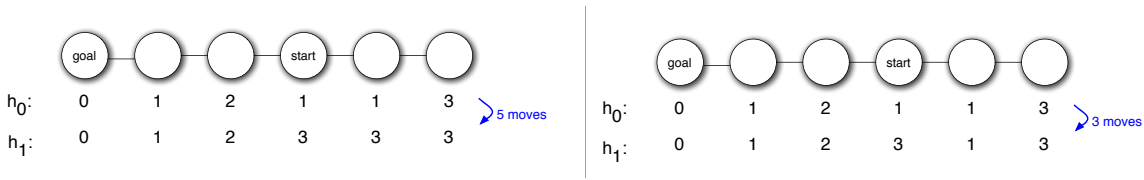

Figure 4: Initial and final heuristics of LRTA* with a lookahead of one (left) and two (right).

Both LRTA* with a lookahead of one and two converge to their final heuristics in one trial. However, the additional ply of lookahead takes advantage of the two extra heuristic values and, as a result, reduces the execution cost from 5 to 3 moves. On the other hand, the planning cost of each move increases from 2 to 4 nodes.

In general, more planning per action "compensates" for inaccuracies in the heuristic function and allows one to select better moves. Table 2 demonstrates this effect averaged over the 1000 pathfinding problems. The reduction of execution cost due to deeper lookahead becomes more pronounced in larger problems (Figure 5).

Lookahead search in LRTA*-like algorithms has received considerable attention in the literature. Russell and Wefald (1991) and Koenig (2004) analyzed selection of the lookahead depth optimal in terms of the *total* cost (i.e., a weighted combination of the planning and execution costs). Bulitko,

Table 2: Effects of deeper lookahead in LRTA*.

| Lookahead | First move lag | Execution convergence cost |
|-----------|----------------|----------------------------|
| 1 | 0.05 ms | 9346 |
| 3 | 0.26 ms | 7795 |
| 5 | 0.38 ms | 6559 |
| 7 | 0.68 ms | 5405 |
| 9 | 0.92 ms | 4423 |

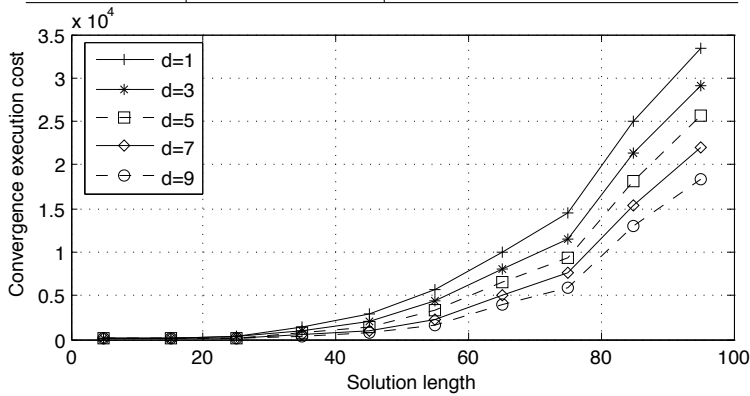

Figure 5: Convergence of LRTA* with different lookaheads as problems scale up.

Li, Greiner, and Levner (2003) and Bulitko (2003) examined pathological cases of deeper lookahead increasing *both* execution and planning costs.

### 6.3 Extension 2: Heuristic Weight

Scalability of LRTA* to large problems is determined by two of its attributes. First, it uses the initial "optimistic" heuristic function for the unknown areas of the search space and more informed, "realistic" function for the explored areas. This results in the so-called "optimism in the face of uncertainty" where the agent eagerly explores novel parts of the space. Consequently, the convergence time is lengthened and the solution quality oscillates significantly over consecutive trials (Shimbo & Ishida, 2003). Second, the complete convergence process can be intractable in terms of running time and the amount of memory required to store the heuristic because optimal solutions are sought. Indeed, even simple problems can be intractable in terms of finding optimal solutions (e.g., the generalized sliding tile puzzle of Ratner & Warmuth, 1986).

A weighted version of LRTA* was proposed by Shimbo and Ishida (2003) as a systematic way of giving up optimality of the converged solution. The authors argued that optimal solutions are rarely needed in practice, nor can they be achieved in large domains. Consequently, the user must settle for suboptimal but satisficing solutions. Weighted LRTA* is the same algorithm (LRTA*) run with an inadmissible initial heuristic. The inadmissibility is bounded as for every state its $h(s)$ value is required to be upper-bounded by $(1 + \epsilon)h^*(s)$ where $h^*(s)$ is the true distance to goal. The resulting $\epsilon$-admissibility leads to $\epsilon$-optimality of the solution to which such an $\epsilon$-LRTA* converges. By giving up optimality in the $\epsilon$-controlled fashion, the algorithm gains faster convergence, smaller memory requirements, and higher stability. Note that 0-LRTA* is equivalent to the original LRTA*.

Shimbo and Ishida (2003) used Manhattan distance scaled by $(1 + \epsilon)$ as the initial heuristic $h_0$. The underlying idea can be illustrated in a five-state domain (Figure 6). On the left, the initial heuristic is listed immediately under the five states (shown as circles). Each action (shown with an edge) has the execution cost of 1. Every trial of LRTA* will involve four moves and will update the heuristic values. The values after each trial are listed as successive rows. After 16 moves (4 trials), LRTA* will converge to the perfect heuristic. On the right, we start with the same heuristic multiplied by $(1 + \epsilon) = 2$. As shown there, 1-LRTA* takes only one trial (i.e., 4 moves) to converge.

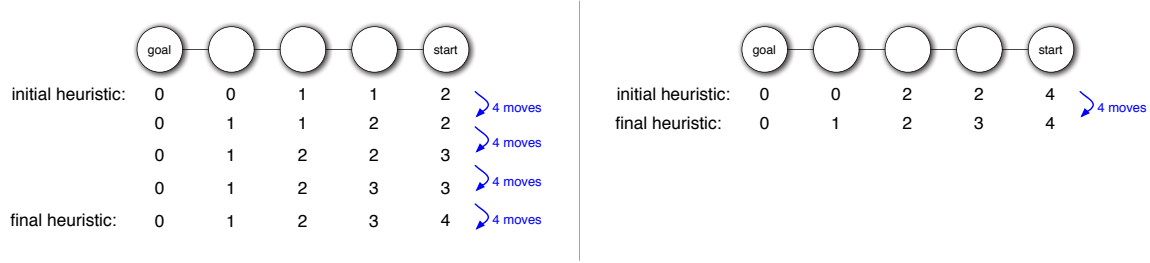

Figure 6: Heuristic values over successive trials of: LRTA* (left) and 1-LRTA* (right).

In general, by scaling the initial (admissible) heuristic by $(1 + \epsilon)$, the amount of underestimation of an admissible initial heuristic relative to the perfect heuristic is reduced in many states. Therefore, the number of updates (line 6 in LRTA*, Figure 2) needed to increase the initial heuristic value of a state to its final value is reduced. Correspondingly, the execution convergence cost is lowered and learning is sped up. Figure 7 illustrates the correlation in the pathfinding domain. The discrepancy is the amount of underestimation in the initial heuristic $(1 + \epsilon)h_0$ averaged over all states of the map at hand ($s \in S$):

$$\operatorname*{avg}_{s_{\text{goal}}, \, s \in S} \frac{h^*(s, s_{\text{goal}}) \dot{-} (1 + \epsilon)h_0(s, s_{\text{goal}})}{h^*(s, s_{\text{goal}})}. \qquad (6.1)$$

Here $\dot{-}$ denotes non-negative subtraction: $a \dot{-} b$ equals $a - b$ if the result is positive and 0 otherwise. As usual, $h^*$ is the perfect heuristic and $h_0$ is the octile distance. The heuristics are taken with respect to the 1000 random goal states ($s_{\text{goal}}$) defined in Section 5.1.

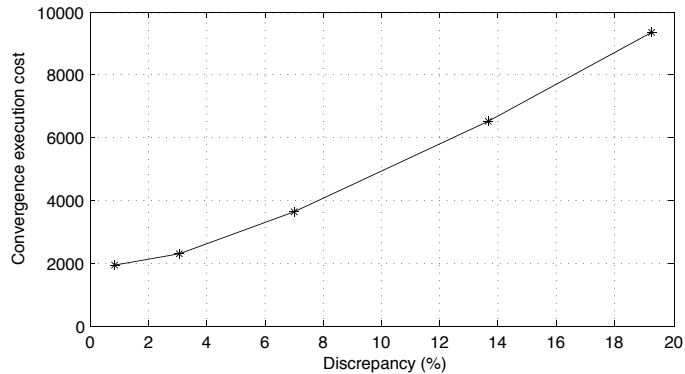

Figure 7: Convergence cost of $\epsilon$-LRTA* vs. the weighted initial heuristic discrepancy. The points on the graph correspond to these values of $(1 + \epsilon)$: 3.3, 2.0, 1.5, 1.1, 1.0.

On the negative side, the scaling is uniform over all states and will sometimes result in the scaled values exceeding the perfect heuristic values. Such inadmissibility can lead to suboptimal solutions as illustrated in Figure 8. Specifically, the initial heuristic is admissible (the values of the five states are shown above the circles in the left part of the figure). All actions (shown as arrows) have the execution cost of 1. Thus, LRTA* converges to the optimal path (i.e., through the top state). On the other hand, scaling the heuristic by $(1 + \epsilon) = 3$ (shown on the right) leads 2-LRTA* to converge on a longer suboptimal path (through the bottom two states).

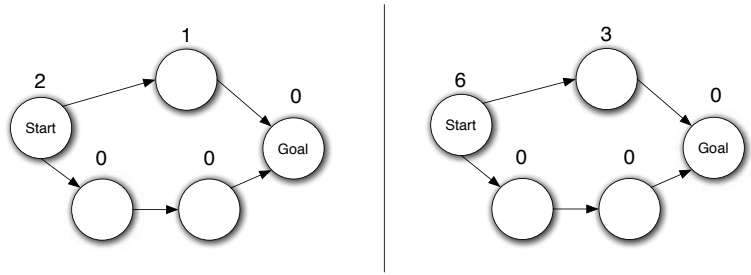

Figure 8: 2-LRTA* converges to a suboptimal path.

In summary, scaling an initial (admissible) heuristic by $1 + \epsilon$ tends to reduce the underestimation error of the heuristic thereby speeding up convergence. As the weighted heuristic becomes progressively more overestimating, suboptimality of the final solutions increases and, eventually, overall convergence slows down (due to less informative weighted heuristic). Table 3 illustrates the trends in the pathfinding domain. Once again, the effect is more pronounced for larger problems (Figure 9).

Note that a similar effect has been observed with weighted A* search wherein a weight $w$ is put on the heuristic $h$. When $w$ exceeds 1, fewer nodes are expanded at the cost of suboptimal solutions (Korf, 1993). Table 4 lists the results of running weighted A* on the pathfinding problems used in Table 3.

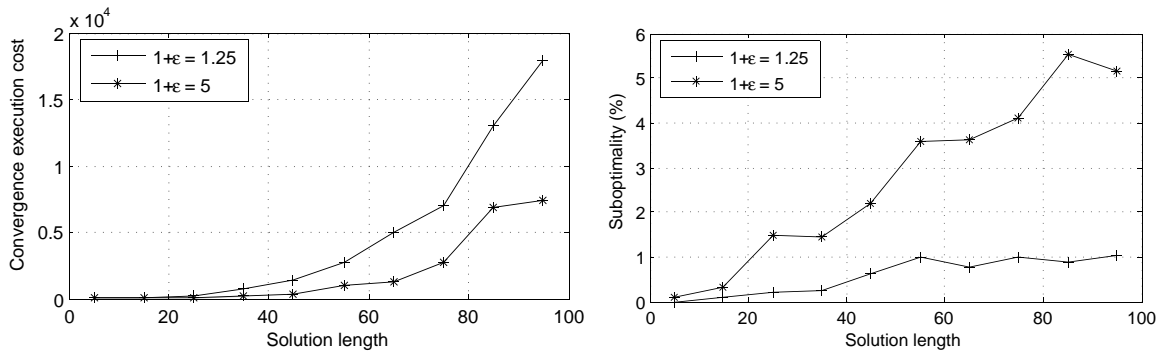

Figure 9: Impact of heuristic weight as pathfinding problems scale up.

## 6.4 Extension 3: Backtracking

Both LRTA* and $\epsilon$-LRTA* learn by updating their heuristic function while advancing in the state space. SLA* (Shue & Zamani, 1993) introduced a backtracking mechanism. That is, upon making an update to heuristic value of the current state, the agent backtracks to its previous state. This provides the agent with two opportunities: (i) to update the heuristic value of the previous state and (ii) possibly select a different action in the previous state. An alternative scheme would be to update heuristic values of previously visited states in memory (i.e., without physically moving the agent there). This is the approach exploited, for example, by temporal difference methods with eligibility traces (Watkins, 1989; Sutton & Barto, 1998). It does not give the agent opportunity (ii) above. Naturally, all backtracking moves are counted in execution cost.

Table 3: Effects of scaling the initial heuristic in $\epsilon$-LRTA* .

| Heuristic weight $(1 + \epsilon)$ | Execution convergence cost | Suboptimality |
|---|---|---|
| 10.0 | 2832 | 3.17% |
| 5.0 | 2002 | 2.76% |
| 3.3 | 1908 | 2.20% |
| 2.0 | 2271 | 1.52% |
| 1.5 | 3639 | 0.97% |
| 1.1 | 6520 | 0.25% |
| 1.0 (LRTA*) | 9346 | 0.00% |

Table 4: Effects of scaling the initial heuristic in weighted A*.

| Heuristic weight | Planning convergence cost | Suboptimality |
|---|---|---|
| 10.0 | 1443 | 3.48% |
| 5.0 | 1568 | 3.08% |
| 3.3 | 1732 | 2.37% |
| 2.0 | 2074 | 1.69% |
| 1.5 | 3725 | 1.16% |
| 1.1 | 5220 | 0.30% |
| 1.0 (A*) | 10590 | 0.00% |

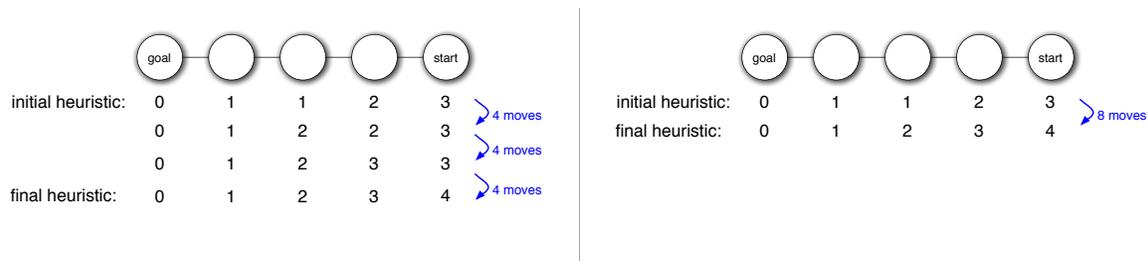

Figure 10: Heuristic values over successive trials of: LRTA* (left) and SLA* (right).

The underlying intuition can be illustrated with a simple example in Figure 10. Once again, consider a one-dimensional five-state domain. Each action has an execution cost of one. The initial heuristic is accurate in the left two states and one lower in the right three states. On each trial, LRTA* will raise the heuristic value of a single state. Therefore, three trials (12 moves) are needed to make the heuristic perfect. SLA*, on the other hand, gets to the middle state in 2 moves, updates its value from 1 to 2, backtracks to the second state from the right, increases its value from 2 to 3, backtracks to the right most state, and increases its value from 3 to 4. The agent will then take 4 moves towards the goal, following the now perfect heuristic. As a result, the first trial is longer (8 vs. 4 moves) but the overall number of moves until convergence is reduced (from 12 to 8).

The SLA* algorithm is nearly identical to LRTA* with a lookahead of one (Figure 2). The only difference is that upon increasing the state value in line 6, the agent takes an action leading to its previous state and not the state $s'$ as in line 7 of LRTA*. If the previous state does not exist (i.e., the algorithm reached the start state) then no action is taken but the heuristic value is still updated.

SLA* backtracks *every time* its heuristic is updated (i.e., learning took place). This causes very substantial execution cost of the first trial. In fact, as we will prove in Theorem 7.8, *all* learning occurs during the first trial. In larger problems, the user may want a suboptimal solution before the convergence process completes. To address this problem, SLA*T, introduced in (Shue, Li, & Zamani, 2001), gives the user control over the amount of learning to be done per trial. Namely, SLA*T backtracks only after it exhausts a user-specified *learning quota*.

There are two primary effects of backtracking on the metrics we consider in this paper. First, larger values of learning quota decrease the execution cost on each trial as less backtracking occurs. In the extreme, the learning quota can be set to infinity, transforming SLA*T into backtracking-free LRTA* with a lookahead of one. On the other end of the spectrum (with a learning quota of zero), SLA*T always backtracks upon updating a heuristic value and becomes equivalent to SLA*. More backtracking tends to speed up the convergence process (i.e., decrease the convergence execution cost) as fewer trials are needed to converge (recall the example in Figure 10). Empirically, the first effect of backtracking is most clearly seen in the 8-puzzle (Table 5) as well as the pathfinding task with small maps (Table 6). It is still an open question why the trend is not observed on larger pathfinding tasks (Table 7).

Table 5: Effects of backtracking in the 8-puzzle.

| Learning quota | First trial execution cost | Convergence execution cost | Memory required |
|---|---|---|---|
| 0 (SLA*) | 1846 | 2321 | 728 |
| 7 | 424 | 52766 | 18751 |
| 13 | 375 | 57816 | 20594 |
| ∞ (LRTA*) | 371 | 58002 | 25206 |

Table 6: Effects of backtracking in the pathfinding domain with small maps.

| Learning quota | First trial execution cost | Convergence execution cost |
|---|---|---|
| 0 (SLA*) | 434 | 457 |
| 10 | 413 | 487 |
| 50 | 398 | 592 |
| 1000 | 390 | 810 |
| ∞ (LRTA*) | 235 | 935 |

The second effect of backtracking is reduction in the amount of memory required to store the learned heuristic values.[7] During backtracking, the agent tends to update heuristic values in previously visited states. Such updates do not require additional memory to be allocated. The memory-saving effect is seen across all problem sizes in Table 5 and Table 7.

## 6.5 Combining the Extensions

The previous sections introduced three extensions of the base algorithm LRTA*. Table 8 lists six algorithms that use the extensions. The first entry summarizes the arbitrary-depth LRTA* which uses the mini-min backup rule to propagate the heuristic values from the search frontier to the interior states in the lookahead tree. It executes one action and converges to an optimal solution. No backtracking is used. Weighted LRTA* ($\epsilon$-LRTA* ) converges to an $\epsilon$-optimal solution but is identical to LRTA* in all other columns of the table. SLA* and SLA*T are myopic versions of LRTA* but use backtracking: unlimited or controlled by the learning quota $T$ respectively. Backtracking was independently introduced in another algorithm, called $\gamma$-Trap (Bulitko, 2004), which also used a heuristic weight parameter $\gamma = 1/(1 + \epsilon)$ to the same effect as weighted LRTA*. Additionally, $\gamma$-Trap employed a more aggressive heuristic update rule, denoted by "max of min" in the table, and used heuristic values of all states (as opposed to the frontier of the lookahead search only). Instead of taking a single action between the lookahead search episodes, it applied $d$ actions to amortize the planning cost.

---

7. Recall that the initial heuristic values are not stored in the table but come from an effectively computable function such as the Manhattan distance.

Table 7: Effects of backtracking in the pathfinding domain with large maps.

| Learning quota | Convergence execution cost | Memory required |
|---|---|---|
| 0 (SLA*) | 9741 | 174 |
| 10 | 10315 | 178 |
| 100 | 11137 | 207 |
| 1000 | 10221 | 258 |
| 10000 | 9471 | 299 |
| $\infty$ (LRTA*) | 9346 | 303 |

Table 8: Real-time heuristic methods compared along several dimensions.

| Algorithm | Lookahead | Learning rule | Execution | Optimality | Backtracking |
|---|---|---|---|---|---|
| LRTA* | arbitrary | mini-min, frontier | one action | optimal | none |
| $\epsilon$-LRTA* | arbitrary | mini-min, frontier | one action | $\epsilon$-optimal | none |
| SLA* | one ply | mini-min, frontier | one action | optimal | always |
| SLA*T | one ply | mini-min, frontier | one action | optimal | controlled by $T$ |
| $\gamma$-Trap | arbitrary | max of min, all states | $d$ actions | $\gamma$-optimal | always |
| **LRTS** | **arbitrary** | **max of min, all states** | **$d$ actions** | **$\gamma$-optimal** | **controlled by $T$** |

We will now combine the three extensions into a single algorithm called Learning Real-Time Search or LRTS. It is shown in the last row of the table and, in a simplified fashion, in Figure 11. The detailed pseudo-code necessary to re-implement the algorithm as well as to follow the theorem proofs is found in Figure 20, Appendix A. We will now step through the operation of LRTS and explain the control parameters.

In line 1, the trial is initialized by setting the current state $s$ of the agent to the start state $s_{\text{start}}$ and resetting the amount of learning done on the trial ($u$). As long as the goal state $s_{\text{goal}}$ is not reached, the LRTS agent interleaves planning (lines 3 and 4), learning (line 5), and execution (lines 6-11). During the planning state, LRTS uses the model of its environment (implicitly updated when new states comes into the agent's radius of visibility) to generate all child states of the current state up to $d$ moves away. On each lookahead level ($i = 1, 2, \ldots, d$), LRTS finds the most promising state ($s_i$) that minimizes the weighted estimate of the total distance:

$$s_i = \arg \min_{s' \in S(s,i)} \left( \gamma \cdot g(s') + h(s') \right), \tag{6.2}$$

where $g(s')$ is the shortest distance from the current state $s$ to its child $s'$ and $h(s')$ is the heuristic estimate of the distance from the child state $s'$ to the goal state $s_{\text{goal}}$. Throughout this paper we will be breaking ties in a systematic fashion as detailed later in the paper. Note that the distance $g(s')$ from the current state $s$ to the child state $s'$ is weighted by the control parameter $\gamma \in (0, 1]$. The same behavior can be obtained by multiplying the initial heuristic by $1 + \epsilon = 1/\gamma$ in $\epsilon$-LRTA* except that it makes the initial heuristic inadmissible (Theorem 7.2).

---

**LRTS**$(d, \gamma, T)$

  1  initialize: $s \leftarrow s_{\text{start}}$, $u \leftarrow 0$
  2  **while** $s \notin S_g$ **do**
  3      generate children $i$ moves away, $i = 1 \dots d$
  4      on level $i$, find the state $s_i$ with the lowest $f = \gamma \cdot g + h$
  5      update $h(s)$ to $\max_{1 \leq i \leq d} f(s_i)$ if it is greater
  6      increase the amount of learning $u$ by $|\Delta h|$
  7      **if** $u \leq T$ **then**
  8          execute $d$ moves to get to the state $s_d$
  9      **else**
 10          execute $d$ moves to backtrack to the previous state, set $u = T$
 11      **end if**
 12  **end while**

---

Figure 11: LRTS algorithm combines the three extensions of LRTA*.

The lookahead search conducted in line 5 has the complexity of $O(b^d)$ if no child state is reached more than once and $b$ is the branching factor. On maps, however, the complexity is much reduced to $O(d^n)$ where $n$ is the dimensionality of the map. Thus, in both the path-planning and the sensor network routing task, the lookahead has a complexity of $O(d^2)$.

Like $\gamma$-Trap , LRTS uses the so-called "max of min" heuristic update (or learning) rule in line 5. Specifically, if the initial heuristic is admissible, then the agent will want to increase it as aggressively as possible while keeping it admissible. Lemma 7.1 will prove that setting the heuristic value $h$ of the current state $s$ to the *maximum* of minima found on each ply preserves the admissibility of $h$ at all times.

In line 6 the amount of increase to the heuristic value of the current state ($h(s)$) is then added to the cumulative amount of learning done so far on the trial ($u$). If the new amount of learning does not exceed the learning quota $T$ then the agent will execute $d$ moves to the most promising frontier state $s_d$ seen during the lookahead (line 8). The moves are added to the path so that they can be "undone" later on during backtracking. If the new amount of learning exceeds the learning quota $T$ then the agent will backtrack to its previous state (line 10). By doing so, it will "undo" the last $d$ actions. The last $d$ actions are thus removed from the path traveled so far on the trial. Note that the backtracking actions count in execution cost.

We now encourage the reader to revisit the hand-traceable examples found in Sections 6.2, 6.3, 6.4 as they show LRTS operation in simple domains. In the weighting example (Section 6.3), $\gamma$ should be set to $1/(1 + \epsilon)$ to enable LRTS to behave as $\epsilon$-LRTA* . This concludes the presentation of the LRTS algorithm and we will now highlight its properties with theoretical and empirical results.

## 7. Theoretical Results

The LRTS framework includes LRTA*, weighted LRTA*, SLA*, and $\gamma$-Trap as special cases (Table 8). This fact is presented formally in the following theorems. All proofs are found in Appendix B. Note that while the following assumptions are not mandatory for real-time search, they are needed for the proofs. Specifically, the search space is assumed to be deterministic and stationary. Additionally, the agent's current state is affected only by the agent's actions. Finally, the agent possesses a sufficient knowledge of the environment to conduct a local-search (i.e., lookahead) of depth $d$ in its current state.

**Theorem 7.1** LRTS($d = 1, \gamma = 1, T = \infty$) is equivalent to LRTA* with a lookahead of one.

**Theorem 7.2** LRTS($d = 1, \gamma = \frac{1}{1+\epsilon}, T = \infty$) initialized with an admissible heuristic $h_0$ is equivalent to $\epsilon$-LRTA* initialized with $(1 + \epsilon)h_0$.

**Theorem 7.3** LRTS($d = 1, \gamma = 1, T = 0$) is equivalent to SLA*.

**Theorem 7.4** LRTS($d, \gamma, T = 0$) is equivalent to $\gamma$-Trap($d, \gamma$).

In the past, convergence and completeness have been proven for the special cases (LRTA*, $\epsilon$-LRTA* , and SLA*). We will now prove these properties for any valid values of the lookahead $d \in \mathbb{N}, d \geq 1$, the heuristic weight $\gamma \in \mathbb{R}, \gamma \in (0, 1]$, and the learning quota $T \in \mathbb{R} \cup \{\infty\}, T \geq 0$.

**Definition 7.1** A tie-breaking scheme is employed when two or more states have equal $f = g + h$ values. Among candidate states with equal $f$-values, *random tie-breaking* selects one at random. This scheme was used in (Shimbo & Ishida, 2003). *Systematic tie-breaking* used by Furcy and Koenig (2000) uses a fixed action ordering per current state. The orderings are, however, generated randomly for each search problem. Finally, *fixed tie-breaking* always uses the same action ordering. We use the latter scheme for all our experiments as it is the simplest.

**Lemma 7.1 (Admissibility)** For any valid $T, \gamma, d$ and an admissible initial heuristic $h_0$, the heuristic function $h$ is maintained admissible at all times during LRTS($d, \gamma, T$) search.

**Theorem 7.5 (Completeness)** For any valid $T, \gamma, d$ and an admissible initial heuristic $h_0$, LRTS($d, \gamma, T$) arrives at a goal state on every trial.

**Theorem 7.6 (Convergence)** For any valid $T, \gamma, d$ and an admissible initial heuristic $h_0$, LRTS($d, \gamma, T$) with systematic or fixed tie-breaking converges to its final solution after a finite number of trials. It makes zero updates to its heuristic function on the final trial. Each subsequent trial will be identical to the final trial.

**Theorem 7.7 (Suboptimality)** For any valid $T, \gamma, d$ and an admissible initial heuristic $h_0$, the execution cost of the final trial of LRTS($d, \gamma, T$) is upper-bounded by $\frac{h^*(s_0)}{\gamma}$. In other words, the suboptimality of the final solution is no worse than $\frac{1}{\gamma} - 1$.

A non-trivial upper-bound of $\frac{h^*(s_0)+T}{\gamma}$ can be imposed on the solution produced by LRTS on the *first* trial. This requires, however, a minor extension of the LRTS algorithm. Thus, we list the extension and the theorem in Figure 21, Appendix A and Theorem B.1, Appendix B respectively.

**Theorem 7.8 (One trial convergence)** For any valid $\gamma, d$ and an admissible initial heuristic $h_0$, the second trial of LRTS($T = 0$) with systematic or fixed tie-breaking is guaranteed to be final. Thus, all learning and exploration are done on the first trial.

## 8. Combinations of Parameters

We have introduced a three-parameter algorithm unifying several previously proposed extensions to the base LRTA*. We have also demonstrated theoretically and empirically the influence of each parameter *independently* of the other two parameters in Sections 6.2, 6.3, and 6.4. The summary of the influences is found in Table 9. An up-arrow/down-arrow means an increase/decrease in the metric. Notation $a \rightarrow p \rightarrow b$ indicates that the parameter $p$ increases from $a$ to $b$. To illustrate: ↓ at the intersection of the row labeled "convergence execution cost" and the column labeled "$1 \rightarrow d \rightarrow \infty$" states that the convergence execution cost decreases as $d$ goes up.

Table 9: Summary of influences of LRTS parameters in the pathfinding domain.

| metric / parameter | $1 \rightarrow d \rightarrow \infty$ | $0 \rightarrow \gamma \rightarrow 1$ | $0 \rightarrow T \rightarrow \infty$ |
|---|---|---|---|
| convergence execution cost | ↓ | ↓ for small $\gamma$, ↑ for large $\gamma$ | ↑ for small $T$, ↓ for large $T$ |
| planning cost per move | ↑ | no influence | no influence |
| convergence memory | ↓ | ↑ | ↑ |
| first-move lag | ↑ | no influence | no influence |
| suboptimality of final solution | ↓ for small $\gamma$ | ↓ | no influence |

### 8.1 Interaction among parameters

In this section we consider effects of parameter *combinations*. Figure 12 shows the impact of the lookahead depth $d$ as a function of the heuristic weight $\gamma$ and the learning quota $T$. Specifically, for each value of $\gamma$ and $T$, the left plot shows the difference between the convergence execution costs for lookahead values of $d = 1$ and $d = 10$:

$$\text{impact-of-d}(\gamma, T) = \text{execution-cost}(d = 1, \gamma, T) - \text{execution-cost}(d = 10, \gamma, T). \quad (8.1)$$

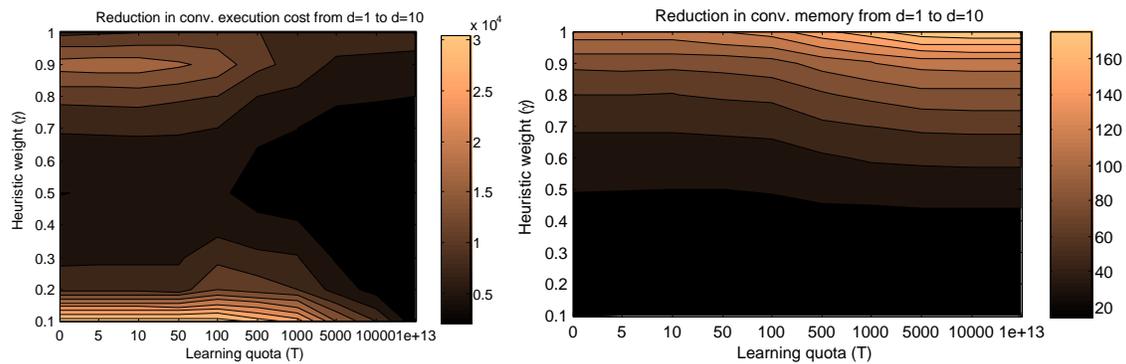

Figure 12: Impact of lookahead $d$ for different values of heuristic weight $\gamma$ and learning quota $T$.

Brighter shades indicate higher impact of increasing $d$ from 1 to 10. Likewise, the right plot demonstrates the impact of $d$ on the memory required for convergence. Brighter shades indicate stronger impact of $d$ (i.e., larger differences). Each point is averaged over the 1000 convergence runs in the pathfinding domain. We observe that larger values of learning quota and higher values of heuristic weight magnify the reduction of memory due to increasing lookahead depth. Note that the plot does not show the memory requirements for different values of $\gamma$ and $T$ but rather the *reduction* in the memory requirements when $d$ increases from 1 to 10.

Figure 13 shows the impact of $\gamma$ as a function of $d$ and $T$ under the same conditions. Learning quota does not seem to affect the impact of $\gamma$ on suboptimality of the final solution. However, deeper lookahead does make $\gamma$ less effective in that respect. We believe it is because deeper lookahead compensates for more inaccuracies in the heuristic function. Therefore, weighting the heuristic with $\gamma$ makes less difference.

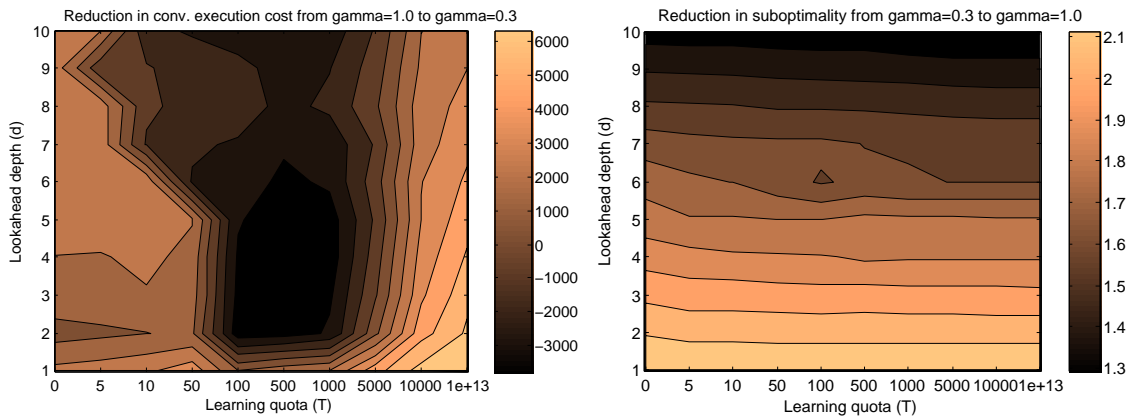

Figure 13: Impact of heuristic weight $\gamma$ for different values of lookahead $d$ and learning quota $T$.

Figure 14 shows the impact of $T$ as a function of $d$ and $\gamma$ on the same 1000 pathfinding problems. Reducing the learning quota $T$ from 500 to 0 results in an increased amount of backtracking. As discussed previously, more backtracking tends to save memory and speed up convergence. The figure shows both trends and demonstrates that $d$ and $\gamma$ affect the impact of backtracking. The left plot shows that more backtracking is most effective in speeding up convergence for lower heuristic weights ($\gamma = 0.1$). On the contrary, the right plot indicates that higher heuristic weights (close to $\gamma = 1$) make backtracking most effective in reducing the memory required for convergence.

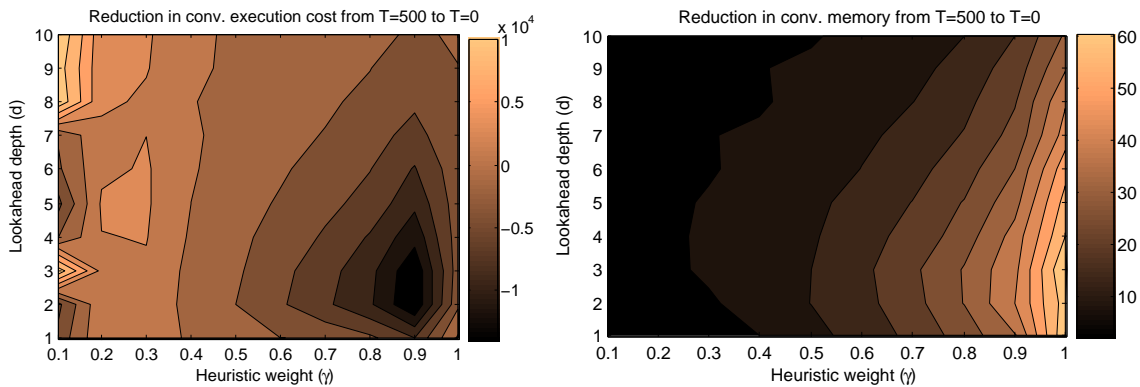

Figure 14: Impact of learning quota $T$ for different values of heuristic weight $\gamma$ and lookahead $d$.

Overall, the influences are non-trivial and, according to our experiments, domain-specific. While some can be explained readily, others appear to be a result of the interaction of the three parameters with the structure of the problem space. Consequently, selection of optimal parameters for a specific performance metric and a concrete domain is presently a matter of trial and error. While manual tuning of control parameters is a typical scenario in Artificial Intelligence, future research will investigate automatic parameter adjustment.

## 8.2 Minimizing performance measures

In this section, we list parameter combinations optimizing each of the performance measures and explain the underlying rationale. The best combinations of parameters are sought in the following parameter space: lookahead $d \in \{1, 2, \ldots, 10\}$, heuristic weight $\gamma \in \{0.1, 0.2, \ldots, 1.0\}$, and the learning quota $T \in \{0, 5, 10, 50, 100, 500, 1000, 5000, 10000, \infty\}$. For each of the one thousand combinations of the parameters, we ran LRTS until convergence on the test suite of 1000 pathfinding problems described in Section 5.1. The results are shown in Table 10 and explained below.

Table 10: LRTS control parameters minimizing performance metrics.

| Metric | Lookahead $d$ | Heuristic weight $\gamma$ | Learning quota $T$ |
|---|---|---|---|
| execution convergence cost | 10 | 0.3 | $\infty$ |
| planning convergence cost | 1 | 0.3 | $\infty$ |
| planning cost per move | 1 | any | any |
| total convergence cost (speed $< 185.638$) | 1 | 0.3 | $\infty$ |
| total convergence cost (speed $\geq 185.638$) | 10 | 0.3 | $\infty$ |
| suboptimality | any | 1.0 | any |
| memory | 10 | 0.3 | 0 |

**Execution convergence cost** is lowest when the quality of individual moves is the highest. Thus, deeper lookahead leading to more planning per move is favored. High values of $\gamma$ are undesirable as LRTS seeks near-optimal solutions and takes longer to converge. Very low values of $\gamma$ are also undesirable as the weighted initial heuristic becomes less and less informative (Table 3). Hence, the best value lays somewhere in the middle (0.3). Finally, no backtracking is preferred as then convergence is the fastest on the larger maps used in this experiment (Table 7).

**Planning convergence cost** is defined as the total planning cost of all moves until convergence is achieved. Deeper lookahead leads to faster convergence but quadratically increases the planning cost of every move (cf., Section 6.5). Hence, overall, shallow lookahead is preferred. This also explains the optimal parameter combination for minimizing **planning cost per move** which in LRTS depends on $d$ only.

**Total convergence cost.** Here we minimize a weighted sum of the planning and execution costs. This is a common objective previously studied in (Russell & Wefald, 1991; Koenig, 2004). We adopt the settings of the latter and bring the two metrics (the execution cost and the planning cost) to a single scalar expressed in the number of states. Namely, our total convergence cost is defined as the convergence execution cost multiplied by the planning speed plus the total planning cost (Section 4). For instance, if the agent converged in 1000 units of execution cost, considered a total of 5000 states while planning (i.e., an average of five states per unit of execution cost), and the planning speed is 200 (i.e., 200 states can be considered in planning while one unit of travel is executed), then the total convergence cost is $1000 \times 200 + 5000 = 205000$, clearly dominated by the execution component. More planning per move tends to decrease the convergence execution cost and thus is preferred when the planning speed is high (i.e., planning is cheap relative to execution and the execution cost dominates the total cost). Conversely, less planning per move is preferred when the planning speed is low. Both Russell and Wefald (1991) and Koenig (2004) used this reasoning to select the optimal lookahead depth in their LRTA*-based real-time search algorithms. In the pathfinding domain, the best values of $\gamma$ and $T$ for fastest convergence are 0.3 and $\infty$ as discussed above. The best value of lookahead depends on the planning speed: deeper lookahead (10 vs. 1) is preferred when the planning speed is sufficiently high and more planning per move can be afforded.

**Suboptimality** is zero with $\gamma = 1$ regardless of $d$ and $T$.

**Memory.** The optimal parameter combination $(10, 0.3, 0)$ is consistent with the previous analysis: more backtracking ($T = 0$) reduces memory consumption, lower heuristic weight $\gamma$ leads to less exploration (since LRTS is satisfied with suboptimal solutions), and since heuristic values are stored $d$ actions apart, higher lookahead is preferred.

Adjusting LRTS control parameters allows the agent to achieve convergence where it has been previously impossible due to memory limitations. In the following, we adopt the settings of Shimbo and Ishida (2003) and Bulitko (2004) and consider one hundred 15-puzzles first published by Korf (1985). It is known that FALCONS and LRTA* with a lookahead of one are unable to converge to optimal solutions on *any* of the 100 puzzles with memory limited to 40 million states (Shimbo & Ishida, 2003). On the other hand, convergence to suboptimal solutions can be achieved with both $\epsilon$-LRTA* (Shimbo & Ishida, 2003) and $\gamma$-Trap (Bulitko, 2004).

Table 11: Convergence on Korf's one hundred 15-puzzles in four million states of memory.

| Algorithm | $d$ | $T$ | $\max \gamma$ | Suboptimality |
|-----------|-----|-----|---------------|---------------|
| FALCONS | - | - | - | no convergence on any problem |
| LRTA* | - | - | - | no convergence on any problem |
| LRTS | 1 | $\infty$ | 0.29 | 50% |
| LRTS | 2 | 0 | 0.6 | 10% |
| **LRTS** | **4** | **0** | **0.7** | **4%** |

We make the task more challenging by reducing the memory limit ten fold from forty to four million states. While we were unable to find a set of $d, T, \gamma$ parameters that allowed LRTS to converge to optimal solutions on all 100 instances, the algorithm did converge to solutions that were, on average, only 4% worse than optimal on *all* 100 puzzles (Table 11). Lower values of $T$ and higher values of $d$ increase the memory cost efficiency and allow to increase $\gamma$ which leads to higher quality solutions.

In summary, adjusting the control parameters of LRTS significantly affects its performance. Our analysis of the individual and combined influence of the parameters on several practically relevant performance metrics gives a practitioner a better intuition for application-specific parameter tuning. Perhaps more importantly, it opens an interesting line of research on automatic selection of optimal parameters on the basis of effectively measurable properties of a domain and the target performance metric.

### 8.3 Application Domain #2: Routing in Sensor Networks

A second application of LRTS considered is routing in *ad hoc* wireless sensor networks. Sensor nodes generally have limited computational power and energy, thus simple, energy efficient routing routines are typically used within sensor networks. Various applications exist, including military (Shang et al., 2003), environmental monitoring (Braginsky & Estrin, 2002) and reconnaissance missions (Yan, He, & Stankovic, 2003). As such, routing in sensor networks is of interest to a large community of users.

The Distance Vector Routing (DVR) algorithm is widely used in network routing (Royer & Toh, 1999). In particular, DVR is employed within the Border Gateway Protocol used to route traffic over the entire Internet. In DVR, each network node maintains a heuristic estimate of its distance to the destination/goal node. We observe that DVR is conceptually very similar to LRTA* with a lookahead of one. Thus, applying LRTS in the network routing domain is essentially equivalent to extending DVR with the heuristic weight $\gamma$ and the backtracking mechanism controlled by the learning quota $T$. Note that there is no analogue of deeper lookahead in DVR since each network node does not learn an explicit model of the network and, consequently, is not aware of any network nodes beyond its immediate neighbors. Hence, $d$ is limited to one.

In order to demonstrate the impact of the two DVR extensions, we chose the setup used in a recent application of real-time heuristic search to network routing (Shang et al., 2003). Specifically, the network sensor nodes are assumed to be dropped from the sky to monitor territory, landing in a random dispersion, and thus necessitating *ad hoc* routing in order to allow for data transmissions. We consider the case of a fixed destination for each transmission, simulating each source node providing information to a central information hub (e.g., a satellite up-link).

Note that LRTA* in the pathfinding domain has access to the heuristic values of its neighbors at no extra cost since all heuristic values are stored in centralized memory. This is not the case in *ad hoc* networks, wherein each node maintains an estimate of the number of hops to the goal node (the $h$ heuristic). A naive application of LRTA* would query the neighbor nodes in order to retrieve their $h$ values. This would, however, substantially increase the amount of network traffic, thereby depleting nodes' energy sources faster and making the network more easily detectable. We adopted the standard solution where each node not only stores its own heuristic value but also the last known heuristic values of its immediate neighbors. Then, whenever a node updates its own heuristic value, it notifies all of its neighbors of the change by broadcasting a short control message. The last major difference from pathfinding is that the nodes do not know either their geographical position or the coordinates of the destination node. Thus, the Euclidian distance is not available and the initial heuristic is initialized to one for all nodes except the destination node where it is set to zero.

In the experiments reported below, $N$ sensor nodes are randomly distributed on an $X \times Y$ two-dimensional grid so that no two nodes occupy the same location. Each node has a fixed transmission radius which determines the number of neighbors a node can broadcast messages to (and whose heuristic values it has to store). To specify a single problem instance, distinct start (source) node and goal (destination or hub) nodes are randomly selected from the $N$ nodes. The start node needs to transmit a data message to the goal node. In doing so, it can send a message to one of its neighbors who will then re-transmit the message to another node, etc. Each transmission between two neighboring nodes incurs an execution cost of one hop. Once the message reaches the destination/goal node, a new trial begins with the same start and goal nodes. Convergence occurs when no heuristic values are updated during the trial.

The objective is to find a shortest route from the start node to the goal node by learning better heuristic values over repeated trials. We take measurements of four metrics paralleling those used in the domain of pathfinding:

**convergence execution cost**  is the total number of hops traveled by messages during convergence;

**first trial execution cost**  is the number of hops traveled on the first trial;

**suboptimality of the final solution**  is the excess of the execution cost of the final trial over the shortest path possible. For instance, if the network converged to a route of 120 hops while the shortest route is 100 hops, the suboptimality is 20%;

**convergence traffic**  is defined as the total volume of information transmitted throughout the network during convergence. Each data message counts as one. Short control messages (used to notify node's neighbors of an update to the node's heuristic function) count as one third.

Interference is not modeled in this simulation. We assume scheduling is performed independently of the routing strategy, and that different schedules will not severely affect the performance of the routing algorithms. Note that, since transmission delay and timings are not modeled, it would be difficult to model interference accurately.

Four batches of networks of 50, 200, 400, and 800 nodes each were considered. The nodes were randomly positioned on a square grid of the dimensions $20 \times 20$, $30 \times 30$, $50 \times 50$, and $80 \times 80$ respectively. Each batch consisted of 100 randomly generated networks. For each network, 25 convergence runs were performed with the parameter values running $\gamma = \{0.1, 0.3, 0.5, 0.7, 1\}$ and $T = \{0, 10, 100, 1000, 100000\}$. In the following, we present the results from the batch of 800-node networks. The same trends were observed in smaller networks as well.

We start out by demonstrating the influence of the two control parameters in isolation. Table 12 shows the influence of the heuristic weight $\gamma$ with no backtracking. As expected, smaller values of $\gamma$ speed up the convergence but increase the suboptimality of the final solution. These results parallel the ones reported in Table 3 for the pathfinding domain.

Table 12: Effects of heuristic weighting in network routing.

| Heuristic weight $\gamma$ | Execution convergence cost | Suboptimality |
|---|---|---|
| 1.0 (DVR) | 8188 | 0% |
| 0.7 | 8188 | 0% |
| 0.5 | 8106 | 0% |
| 0.3 | 7972 | 0.29% |
| 0.1 | 7957 | 0.33% |

Table 13 demonstrates the influence of the learning quota $T$ when the heuristic weight $\gamma$ is set to one. Smaller values of $T$ increase the amount of backtracking and speed up convergence cost but lengthen the first trial. This parallels the results in the 8-puzzle (Table 5) as well as in pathfinding with small maps (Table 6).

We will now consider the impact of parameter combination. The fixed lookahead depth of one makes the two-dimensional parameter space easier to visualize on contour plots. Thus, instead of plotting the *impact* of a parameter on a metric as a function of the other two control parameters, we will now plot the *metric* itself. Figure 15 shows the first trial and convergence execution costs for all combinations of $\gamma$ and $T$. Brighter areas indicates higher costs.

Table 13: Effects of backtracking in network routing.

| Learning quota $T$ | First trial execution cost | Convergence execution cost |
|---|---|---|
| $10^5$ (DVR) | 549 | 8188 |
| $10^3$ | 1188 | 8032 |
| $10^2$ | 4958 | 6455 |
| 10 | 5956 | 6230 |
| 0 | 6117 | 6138 |

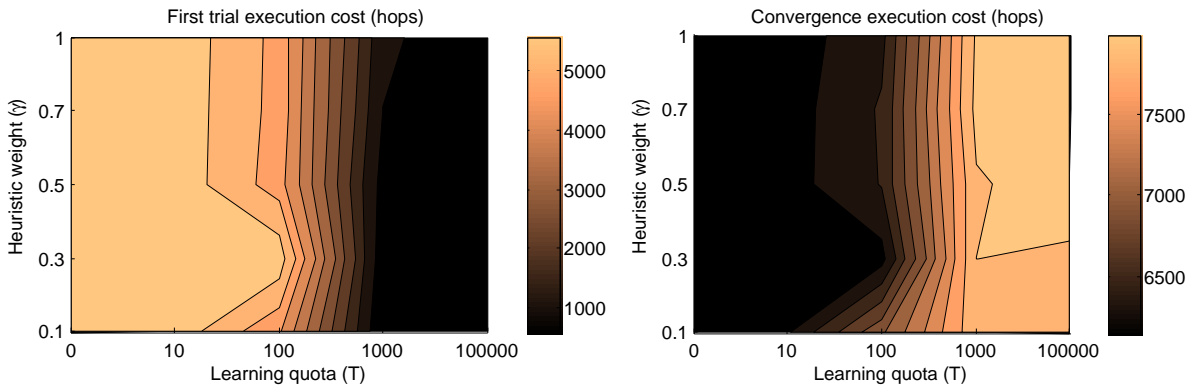

Figure 15: First-trial and convergence execution costs in network routing.

Figure 16 shows the suboptimality of the final solution and the total traffic on a convergence run, averaged over 100 networks of 800 nodes each. Again, brighter areas indicate higher values.

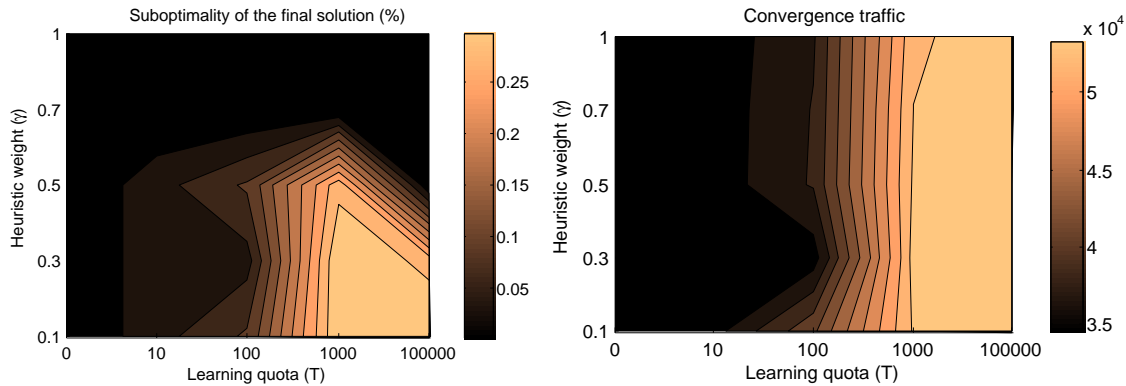

Figure 16: Suboptimality of the final solution and the total traffic in network routing.

In summary, by extending the DVR algorithm with the heuristic weight and the backtracking mechanisms of LRTS, reduction of the convergence execution cost and the total network traffic can be achieved. As a result, near-optimal routes are found faster, at a lower energy consumption. This promising result raises several questions for further research. First, how will additional routing constraints, such as the ones investigated in (Shang et al., 2003), affect the performance of LRTS? Second, how will the benefits of LRTS scale to the case of several messages with the same destina-tion passing through the network *simultaneously*? The latter question also applies to path-planning with multiple memory-sharing units which is a common situation in squad-based games.

## 9. Summary and Future Research

In this paper, we considered simultaneous planning and learning problems such as real-time pathfinding in initially unknown environments as well as routing in *ad hoc* wireless sensor networks. Learning real-time search methods plan by conducting a limited local search and learn by maintaining and refining a state-based heuristic function. Thus, they can react quickly to user's commands and yet converge to better solutions over repeated experiences.

We analyzed three complementary extensions of the base LRTA* algorithm and showed their effects with hand-traceable examples, theoretical bounds, and extensive empirical evaluation in two different domains. We then proposed a simple algorithm, called LRTS, unifying the extensions and studied the problem of choosing optimal parameter combinations.

Current and future work includes extending LRTS with automated state space abstraction mechanisms, function approximation methods for the heuristic, and automated parameter selection techniques. It will also be of interest to investigate the convergence process of a team of LRTS-based agents with shared memory and extend LRTS for moving targets and stochastic/dynamic environments in video games, robotics, and networking.

## Acknowledgements

Input from Valeriy Bulitko, Sven Koenig, Rich Korf, David Furcy, Nathan Sturtevant, Masashi Shimbo, Rob Holte, Douglas Aberdeen, Reza Zamani, Stuart Russell, and Maryia Kazakevich is greatly appreciated. Nathan Sturtevant kindly provided and supported Hierarchical Open Graph simulator. Additionally, Valeriy Bulitko, David Furcy, Sven Koenig, Rich Korf, Scott Thiessen, and Ilya Levner have taken time to proof read an early draft of this paper. Great thanks to JAIR reviewers for working with us on improving the manuscript. We are grateful for the support from NSERC, the University of Alberta, the Alberta Ingenuity Centre for Machine Learning, and Jonathan Schaeffer.

## Appendix A. Detailed Pseudocode

In this section we list pseudocode of the algorithms at the level of detail necessary to implement them as well as to follow the proofs listed in the Appendix B.

---

**LRTA\***

OUTPUT: a series of actions leading from $s_0$ to a goal state

| | |
|---|---|
| 1 | **if** $h$ is undefined **then** set $h$ to $h_0$ |
| 2 | reset the current state: $s \leftarrow s_0$ |
| 3 | **while** $s \notin S_g$ **do** |
| 4 |     generate depth 1 neighborhood $S(s,1) = \{s' \in S \mid \|s,s'\| = 1\}$ |
| 5 |     compute $h'(s) = \min\limits_{s' \in S(s,1)} (\text{dist}(s,s') + h(s'))$ |
| 6 |     **if** $h'(s) > h(s)$ **then** update $h(s) \leftarrow h'(s)$ |
| 7 |     update current state $s \leftarrow \arg\min\limits_{s' \in S(s,1)} (\text{dist}(s,s') + h(s'))$ |

---

Figure 17: LRTA\* algorithm.

---

**SLA\***

OUTPUT: a series of actions leading from $s_0$ to a goal state

| | |
|---|---|
| 1 | **if** $h$ is undefined **then** set $h$ to $h_0$ |
| 2 | reset the current state: $s \leftarrow s_0$ |
| 3 | reset the path traveled so far: $path \leftarrow \emptyset$ |
| 4 | **while** $s \notin S_g$ **do** |
| 5 |     generate depth 1 neighborhood $S(s,1) = \{s' \in S \mid \|s,s'\| = 1\}$ |
| 6 |     compute $h'(s) = \min\limits_{s' \in S(s,1)} (\text{dist}(s,s') + h(s'))$ |
| 7 |     **if** $h'(s) > h(s)$ **then** |
| 8 |         update $h(s) \leftarrow h'(s)$ |
| 9 |         update current state $s \leftarrow pop(path)$ [or remain in state $s$ if $path = \emptyset$] |
| 10 |     **else** |
| 11 |         push $s$ onto the stack $path$ |
| 12 |         update current state $s \leftarrow \arg\min\limits_{s' \in S(s,1)} (\text{dist}(s,s') + h(s'))$ |

---

Figure 18: SLA\* algorithm.

---

$\gamma$-**Trap**$(d, \gamma)$

Input:

    $d$:    lookahead depth

    $\gamma$:    optimality weight

Output: $path$: a series of actions leading from $s_0$ to a goal state

1    **if** $h$ is undefined **then** set $h$ to $h_0$

2    reset the current state: $s \leftarrow s_0$

3    reset the stack: $path \leftarrow \emptyset$

4    reset the cumulative learning amount $u \leftarrow 0$

5    **while** $s \notin S_g$ **do**

6        set $d \leftarrow \min\{d, \min\{k \mid S(s,k) = \emptyset\}\}$

7        reset set $G \leftarrow \emptyset$

8        **for** $k$=1 to $d$ **do**

9            generate depth $k$ neighborhood $S(s,k) = \{s' \in S \mid \|s, s'\| = k\}$

10            **if** $S(s,k) \cap S_g \neq \emptyset$ **then** update $G \leftarrow G \cup \{k\}$

11            compute $f_{\min}(s,k) = \min\limits_{s' \in S(s,k)} (\gamma \cdot \text{dist}(s,s') + h(s'))$

12            compute $s_{\min}(s,k) = \arg\min\limits_{s' \in S(s,k)} (\gamma \cdot \text{dist}(s,s') + h(s'))$

13        **end for**

14        compute $h'(s) \leftarrow \begin{cases} \max\limits_{1 \leq k \leq d} f_{\min}(s,k) & \text{if } G = \emptyset, \\ \max\limits_{1 \leq k \leq \min\{G\}} f_{\min}(s,k) & \text{otherwise.} \end{cases}$

15        **if** $h'(s) \leq h(s)$ **then**

16            push $s$ onto stack $path$

17            update the current state $s \leftarrow \begin{cases} s_{\min}(s,d) & \text{if } G = \emptyset, \\ \arg\min\limits_{k \in G} f_{\min}(s,k) & \text{otherwise.} \end{cases}$

18        **else**

19            update $h(s) \leftarrow h'(s)$

20            backtrack: $s \leftarrow pop(path)$ [or remain in state $s$ if $path = \emptyset$].

21        **end if**

22    **end while**

---

Figure 19: $\gamma$-Trap algorithm.

---

**LRTS**$(d, \gamma, T)$

INPUT:

| | |
|---|---|
| $d'$: | lookahead depth |
| $\gamma$: | optimality weight |
| $T$: | learning threshold |

OUTPUT: $path$: a series of actions leading from $s_0$ to a goal state

1   **if** $h$ is undefined **then** set $h$ to $h_0$
2   reset the current state: $s \leftarrow s_0$
3   reset the stack: $path \leftarrow \emptyset$
4   reset the cumulative learning amount $u \leftarrow 0$
5   **while** $s \notin S_g$ **do**
6       set $d \leftarrow \min\{d', \min\{k \mid S(s, k) = \emptyset\}\}$
7       reset set $G \leftarrow \emptyset$
8       **for** $k$=1 to $d$ **do**
9           generate depth $k$ neighborhood $S(s, k) = \{s' \in S \mid \|s, s'\| = k\}$
10          **if** $S(s, k) \cap S_g \neq \emptyset$ **then** update $G \leftarrow G \cup \{k\}$
11          compute $f_{\min}(s, k) = \min\limits_{s' \in S(s,k)} (\gamma \cdot \text{dist}(s, s') + h(s'))$
12          compute $s_{\min}(s, k) = \arg\min\limits_{s' \in S(s,k)} (\gamma \cdot \text{dist}(s, s') + h(s'))$
13      **end for**
14      compute $h'(s) \leftarrow \begin{cases} \max\limits_{1 \leq k \leq d} f_{\min}(s, k) & \text{if } G = \emptyset, \\ \max\limits_{1 \leq k \leq \min\{G\}} f_{\min}(s, k) & \text{otherwise.} \end{cases}$
15      **if** $h'(s) > h(s)$ **then**
16          compute amount of learning on this move: $\ell \leftarrow h'(s) - h(s)$
17          update $h(s) \leftarrow h'(s)$
18      **else**
19          reset $\ell = 0$
20      **end if**
21      **if** $u + \ell \leq T$ **then**
22          push $s$ onto stack $path$
23          update the current state: $s \leftarrow \begin{cases} s_{\min}(s, d) & \text{if } G = \emptyset, \\ \arg\min\limits_{k \in G} f_{\min}(s, k) & \text{otherwise.} \end{cases}$
24          accumulate learning amount $u \leftarrow u + \ell$
25      **else**
26          backtrack: $s \leftarrow pop(path)$  [or remain in state $s$ if $path = \emptyset$].
27      **end if**
28  **end while**

---

Figure 20: LRTS algorithm.

---

23a          **if** $\exists s' [s = s' \And s' \in path]$ **then**
23b              remove $s'$ and all following states from $path$
23c          **end if**

---

Figure 21: Additional lines facilitating incremental pruning in LRTS. These are to be inserted between lines 23 and 24 in Figure 20.

## Appendix B. Theorem Proofs

**Theorem 7.1** LRTS($d = 1, \gamma = 1, T = \infty$) is equivalent to LRTA* with a lookahead of one.

**Proof.** We will show that in any state $s$ LRTS($d = 1, \gamma = 1, T = \infty$) (i) takes the same action as LRTA* does and (ii) updates the $h$ function in the same way LRTA* does.

(i) In state $s$, LRTS($d = 1, \gamma = 1, T = \infty$) goes to state $s_{\min}(s, 1)$ (line 23, Figure 20) which is equivalent to the state $\arg \min_{s_1 \in S(s,1)} (\text{dist}(s, s_1) + h(s_1))$ to which LRTA* will move (line 7, Figure 17).[8]

(ii) LRTS($d = 1, \gamma = 1, T = \infty$) updates $h(s)$ to $h'(s)$ in line 17 in Figure 20 when $h'(s) > h(s)$ (line 15) which is equivalent to LRTA*'s update under the same condition in line 6, Figure 17. $\square$

**Theorem 7.2** LRTS($d = 1, \gamma = \frac{1}{1+\epsilon}, T = \infty$) initialized with an admissible heuristic $h_0$ is equivalent to $\epsilon$-LRTA* initialized with $(1 + \epsilon)h_0$.

**Proof.** These two algorithms begin with and maintain different heuristic functions. Let $h_t^{\text{LRTS}}$ be the heuristic function LRTS($d = 1, \gamma = \frac{1}{1+\epsilon}, T = \infty$) maintains at iteration $t$.[9] Likewise, let $h_t^{\epsilon\text{-LRTA*}}$ denote the heuristic function $\epsilon$-LRTA* maintains at iteration $t$. We will also denote by $s_t^{\text{LRTS}}$ the current state of LRTS($d = 1, \gamma = \frac{1}{1+\epsilon}, T = \infty$) at iteration $t$. Likewise, $s_t^{\epsilon\text{-LRTA*}}$ is the current state of $\epsilon$-LRTA* at iteration $t$. We show by induction over iteration number $t$ that for $\forall t$:

$$h_t^{\epsilon\text{-LRTA*}} = (1 + \epsilon)h_t^{\text{LRTS}}. \tag{B.1}$$

Base step: since $\epsilon$-LRTA* initialized with $(1 + \epsilon)h_0$ and LRTS($d = 1, \gamma = \frac{1}{1+\epsilon}, T = \infty$) is initialized with $h_0$, equation B.1 trivially holds at $t = 0$.

Inductive step: suppose equation B.1 holds at iteration $t$. We will show that it holds at iteration $t + 1$.

First, we show that both algorithms are bound to be at the same state $s_t$. Suppose not: $s_t^{\text{LRTS}} \neq s_t^{\epsilon\text{-LRTA*}}$. Then since both start in state $s_0$ there must have been the earliest $t_0 < t$ such that $s_{t_0}^{\text{LRTS}} = s_{t_0}^{\epsilon\text{-LRTA*}}$ but $s_{t_0+1}^{\text{LRTS}} \neq s_{t_0+1}^{\epsilon\text{-LRTA*}}$. This means that in state $s_{t_0}$ different actions were taken by LRTS and $\epsilon$-LRTA*. LRTS takes action in line 23 of Figure 20 and would move to state:

$$\arg \min_{s' \in S(s_{t_0},1)} \left( \frac{1}{1+\epsilon} \cdot \text{dist}(s_{t_0}, s') + h_{t_0}^{\text{LRTS}}(s') \right). \tag{B.2}$$

$\epsilon$-LRTA* moves to the following state (line 7 in Figure 17):

$$\arg \min_{s' \in S(s_{t_0},1)} \left( \text{dist}(s_{t_0}, s') + h_{t_0}^{\epsilon\text{-LRTA*}}(s') \right). \tag{B.3}$$

Since $h_{t_0}^{\epsilon\text{-LRTA*}} = (1 + \epsilon)h_{t_0}^{\text{LRTS}}$ (as $t_0 < t$ and equation B.1 holds at time $t$ by the inductive hypothesis), the action in B.2 is the same as in B.3.

---

8. Throughout the paper we assume that tie-breaking is done in a consistent way among all algorithms.

9. One can measure iterations by counting the number of times the **while** condition is executed (line 5 in Figure 20 and line 3 in Figure 17).

The update to $h$ in LRTS($d = 1, \gamma = \frac{1}{1+\epsilon}, T = \infty$) occurs in line 17, Figure 20. Substituting the expression for $h'(s)$ we arrive at:

$$
\begin{align}
h_{t+1}^{\text{LRTS}}(s_t) &= \min_{s' \in S(s_t, 1)} \left( \frac{1}{1+\epsilon} \operatorname{dist}(s_t, s') + h_t^{\text{LRTS}}(s') \right) \tag{B.4} \\
&= \frac{1}{1+\epsilon} \min_{s' \in S(s_t, 1)} \left( \operatorname{dist}(s_t, s') + (1+\epsilon) h_t^{\text{LRTS}}(s') \right) \tag{B.5} \\
&= \frac{1}{1+\epsilon} \min_{s' \in S(s_t, 1)} \left( \operatorname{dist}(s_t, s') + h_t^{\epsilon\text{-LRTA*}}(s') \right) \tag{B.6} \\
&= \frac{1}{1+\epsilon} h_{t+1}^{\epsilon\text{-LRTA*}}(s_t) \tag{B.7}
\end{align}
$$

which concludes the inductive proof. $\square$

**Theorem 7.3** LRTS($d = 1, \gamma = 1, T = 0$) is equivalent to SLA*.

**Proof.** When $T = 0$, condition $u + \ell \leq T$ in line 21 of Figure 20 will hold if and only if $h(s)$ is not updated (i.e., $h'(s) \leq h(s)$ in line 15). If that is the case the forward move executed in line 23 is equivalent to SLA*'s forward move in lines 11, 12 of Figure 18. If $h(s)$ is indeed updated then both LRTS and SLA* will backtrack by executing line 26 in Figure 20 and line 9 in Figure 18 respectively. The proof is concluded with the observation that $h'(s)$ computed by LRTS($d = 1, \gamma = 1, T = 0$) is equivalent to $h'(s)$ computed by SLA* in line 6 of Figure 18. $\square$

**Theorem 7.4** LRTS($d, \gamma, T = 0$) is equivalent to $\gamma$-Trap($d, \gamma$).

**Proof.** Substituting $T = 0$ in line 21 of Figure 20, we effectively obtain $\gamma$-Trap (Figure 19). $\square$

**Lemma 7.1 (Admissibility)** For any valid $T, \gamma, d$ and an admissible initial heuristic $h_0$, the heuristic function $h$ is maintained admissible at all times during LRTS($d, \gamma, T$) search.

**Proof.** We will prove this lemma by induction over the iteration number $t$. At $t = 0$ the statement trivially holds since the initial heuristic $h_0$ is admissible. Suppose $h_t$ is admissible then we will show that $h_{t+1}$ is admissible. Heuristic $h_{t+1}$ can be different from $h_t$ only in state $s_t$ as it is updated in lines 14 and 17 of Figure 20. Combining the expressions together, we obtain:

$$
h_{t+1}(s_t) = \max_{1 \leq k \leq d} \min_{s' \in S(s_t, k)} (\gamma \operatorname{dist}(s_t, s') + h_t(s')), \tag{B.8}
$$

if no goal states have been discovered during the lookahead ($G = \emptyset$) and:

$$
h_{t+1}(s_t) = \min_{k \in G} \min_{s' \in S(s_t, k)} (\gamma \operatorname{dist}(s_t, s') + h_t(s')), \tag{B.9}
$$

otherwise. Suppose the $\max \min$ or $\min \min$ is reached in state $s^\circ$:

$$
h_{t+1}(s_t) = \gamma \operatorname{dist}(s_t, s^\circ) + h_t(s^\circ), s^\circ \in S(s_t, m), 1 \leq m \leq d. \tag{B.10}
$$

The shortest path from state $s_t$ and the closest goal state intersects neighborhood $S(s_t, m)$ of depth $m$. Let us denote state from $S(s_t, m)$ belonging to the shortest path by $s^*$ (Figure 22).

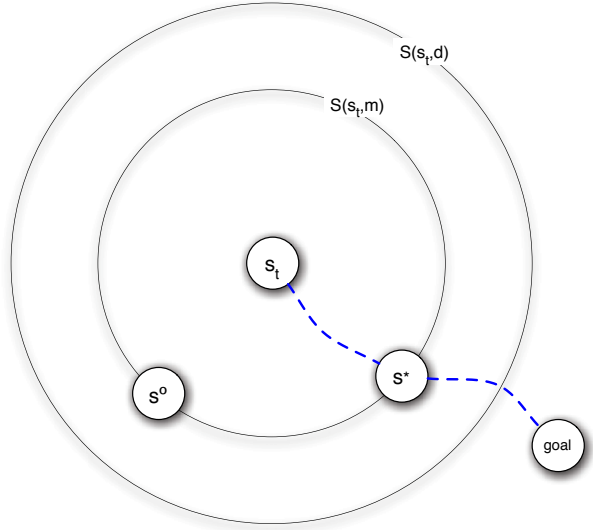

Figure 22: To the proof of admissibility.

Since $h_t$ is admissible in all states, $\gamma \leq 1$, and $s^\circ$ was chosen as $\arg \min\limits_{s \in S(s_t, m)} (\gamma \operatorname{dist}(s_t, s) + h_t(s))$ we conclude that:

$$
\begin{aligned}
h_{t+1}(s_t) &= \gamma \operatorname{dist}(s_t, s^\circ) + h_t(s^\circ) & \text{(B.11)} \\
&\leq \gamma \operatorname{dist}(s_t, s^*) + h_t(s^*) & \text{(B.12)} \\
&\leq \operatorname{dist}(s_t, s^*) + h^*(s^*) & \text{(B.13)} \\
&= h^*(s_t). & \text{(B.14)}
\end{aligned}
$$

Thus, $h_{t+1}$ is an admissible heuristic which concludes the induction. □

**Theorem 7.5 (Completeness)** For any valid $T, \gamma, d$ and an admissible initial heuristic $h_0$, LRTS$(d, \gamma, T)$ arrives at a goal state on every trial.

**Proof.** Since heuristic $h$ is maintained admissible at all times and each update increases its value by a positive amount lower-bounded by a constant, only a finite number of updates can be made on any trial. Let the last update be made in state $s_{u-1}, u \geq 0$ in the sequence of states $s_0, s_1, \ldots$ traversed on the trial. We will show that the sequence of states $s_u, s_{u+1}, \ldots$ is guaranteed to terminate in a goal state. In other words, LRTS cannot loop in the finite state space forever avoiding the goal states. As there are no updates made in $s_u, s_{u+1}, \ldots$, it must hold that[10] for $\forall j \geq u$:

$$
\begin{aligned}
h(s_j) &\geq \max_{1 \leq k \leq d} \min_{s' \in S(s_j, k)} (\gamma \operatorname{dist}(s_j, s') + h(s')), & \text{(B.15)} \\
h(s_j) &\geq \gamma \operatorname{dist}(s_j, s_{j+1}) + h(s_{j+1}), & \text{(B.16)} \\
h(s_j) - h(s_{j+1}) &\geq \gamma \operatorname{dist}(s_j, s_{j+1}). & \text{(B.17)}
\end{aligned}
$$

Since $\gamma > 0$ and for any states $a \neq b \,[\operatorname{dist}(a, b) > 0]$, it follows that $h(s_u) > h(s_{u+1}) > h(s_{u+2}) > \ldots$. As the state space is finite, this series of states must terminate in a goal state with $h(s_n) = 0$. Thus, for any valid parameter values, LRTS ends in a goal state on every trial. □

---

10. Similar reasoning applies to the case of $G \neq \emptyset$ when min min is used instead of max min.

**Theorem 7.6 (Convergence)** For any valid $T, \gamma, d$ and an admissible initial heuristic $h_0$, LRTS$(d, \gamma, T)$ with systematic or fixed tie-breaking converges to its final solution after a finite number of trials. It makes zero updates to its heuristic function on the final trial. Each subsequent trial will be identical to the final trial.

**Proof.** From the proof of Theorem 7.5 it is clear that there will be a trial with zero updates to the heuristic function $h$. We will call this trial the final trial and will demonstrate that each subsequent trial will be identical to the final trial. Suppose not. Then in the final trial $s_0, s_1, \ldots, s_n$ there is the earliest state $s_j$ such that the next trial is different from the final trial in state $s_{j+1}$. This implies that on the subsequent trial LRTS took a different action in state $s_j$ from the action it took in the same state on the final trial. As there are no updates to the heuristic function, the action must have been selected in line 23 of Figure 20. Expanding the expression, we get:[11]

$$s_{j+1} = s_{\min}(s_j, d) = \arg \min_{s' \in S(s_j, d)} (\gamma \operatorname{dist}(s_j, s') + h(s')). \tag{B.18}$$

Since the ties are broken in a systematic or fixed fashion, the choice of state $s_{j+1}$ is unique which contradicts with our assumption on existence of a subsequent trial different from the final trial. $\square$

**Theorem 7.7** For any valid $T, \gamma, d$ and an admissible initial heuristic $h_0$, the converged solution cost[12] of LRTS$(d, \gamma, T)$ is upper-bounded by $\frac{h^*(s_0)}{\gamma}$.

**Proof.** First, let us denote the converged solution cost (i.e., the execution cost of the final trial) by $\widehat{\text{solution}}$. From theorem 7.6 it follows that there will be a final trial with no updates to the heuristic function. Suppose $s_0, \ldots, s_n$ are the states traversed during the final trial. Since no updates are made to the heuristic function, the condition $h'(s_i) \leq h(s_i)$ in line 15 of Figure 20 must have held for all $s_i, i = 0, \ldots, n$. Substituting the expression for $h'(s_i)$ from line 14, we arrive at:[13]

$$\max_{1 \leq k \leq d} \min_{s' \in S(s_i, k)} \left( \gamma \operatorname{dist}(s_i, s') + h(s') \right) \leq h(s_i). \tag{B.19}$$

Since:

$$s_{i+1} = \min_{s' \in S(s_i, d)} \left( \gamma \operatorname{dist}(s_i, s') + h(s') \right) \tag{B.20}$$

we conclude that:

$$\gamma \operatorname{dist}(s_i, s_{i+1}) + h(s_{i+1}) \quad \leq \quad h(s_i) \tag{B.21}$$

$$\gamma \operatorname{dist}(s_i, s_{i+1}) \quad \leq \quad h(s_i) - h(s_{i+1}). \tag{B.22}$$

Summing up these telescoping inequalities for $i = 0, \ldots, n-1$, we derive:

$$\sum_{i=0}^{n-1} \operatorname{dist}(s_i, s_{i+1}) \leq \frac{h(s_0) - h(s_n)}{\gamma}. \tag{B.23}$$

---

11. The case of $G \neq \emptyset$ is handled similarly.
12. Henceforth "converged solution cost" is defined as the execution cost of the final trial.
13. Once again, the case of $G \neq \emptyset$ is handled similarly.

Since $s_n \in S_g$, $h(s_n) = 0$. The sum of distances travelled is the travel cost of the final trial and is the converged solution cost by definition.[14] Thus, we conclude that the converged solution cost is upper bounded by $\frac{h(s_0)}{\gamma} \leq \frac{h^*(s_0)}{\gamma}$. $\square$

**Theorem 7.8** For any valid $\gamma, d$ and an admissible initial heuristic $h_0$, the second trial of LRTS($T = 0$) with systematic or fixed tie-breaking is guaranteed to be final. Thus, all learning and exploration are done on the first trial.

**Proof.** Consider a sequence of states $[s_0, s_1, \ldots, s_n]$ traversed during the first trial. Here $s_0$ is the start state and $s_n$ is the goal state found. Each transition $s_t \rightarrow s_{t+1}, t = 0, \ldots, n-1$ was carried out via a forward move (line 23 in Figure 20) or backtracking move (line 26). During move $t = 0, \ldots, n-1$, the heuristic function may have been changed (line 17) in state $s_t$. We will use $h_t$ to denote the heuristic function before the update in line 17. The update results in $h_{t+1}$. Note that the update changes $h$ in state $s_t$ only so that $h_t(s) = h_{t+1}(s)$ for all $s \neq s_t$. In particular, since $s_{t+1} \neq s_t$[15] we have $h_{t+1}(s_{t+1}) = h_t(s_{t+1})$.

Since $T = 0$, all $n$ moves can be divided in two groups: forward moves that do not change the $h$ function and backtracking moves which do. We will now prune the sequence of states $[s_0, s_1, \ldots, s_n]$ using a backtracking stack. Namely, at the beginning the stack contains $s_0$. Every forward move $0 \leq t \leq n-1$ will add $s_{t+1}$ to the stack. Every backtracking move $0 \leq t \leq n-1$ will remove $s_{t+1}$ from the stack. Carrying this procedure for $t = 0, \ldots, n-1$, we derive the sequence of states: $p' = [s_0, \ldots, s_m]$ leading from the initial state $s_0$ to the goal state $s_m$. Clearly, $m \leq n$.

Observe that since all backtracking moves were removed from the original move sequence, the only moves left are forward moves with no updates to $h$. Thus, let us define $\widehat{h}$-value of any state $s$ visited on the trial as:

$$\widehat{h}(s) = \begin{cases} h(s) & \text{after the most recent backtracking move from state } s; \\ \text{initial } h(s) & \text{if no backtracking move was taken in } s \text{ during the trial.} \end{cases}$$

Since now $T = 0$, for any state $s_i, i = 0, \ldots, m$ there have been no updates in $h$ *after* the last backtracking move from that state (if any). In other words, if $h$ is the final heuristic then $\forall i = 0, \ldots, m \left[ h(s_i) = \widehat{h}(s_i) \right]$. It also means that in every $k$-neighbor ($1 \leq k \leq d$) of each state $s_i \in p'$ there was a state $s'$ with $\widehat{h}(s') + \gamma \operatorname{dist}(s_i, s') \leq \widehat{h}(s_i)$.

We now have to show that no updates to $h$ will be done on the second trial. Suppose this is not the case. Then there must exist the earliest state $s_j \in p', 0 \leq j < m$ such that LRTS on the second trial updated $h(s_j)$ to be greater than $\widehat{h}(s_j)$. As we reasoned above, the final value of $h(s_j)$ on the first trial was $\widehat{h}(s_j)$ and there was no need to increase it. By the assumption, LRTS did increase $h(s_j)$ on the second trial. Consequently, at least one state in one of $s_j$'s neighborhoods was updated during *first trial* (as no updates on the second trial had happened yet).

Formally, $h(s')$, where $s' \in S(s_j, k)$ for some $1 \leq k \leq d$, was increased during the first trial but *after* state $s_j$ was visited in $p'$. Additionally, this means that as a result of this update it must hold that:

$$h_{\text{new}}(s') + k > h(s_j). \tag{B.24}$$

---

14. It is easy to check that there the sequence of states on the final trial has no repetitions. Therefore, $\widehat{\text{solution}} = \sum_{i=0}^{n-1} \operatorname{dist}(s_i, s_{i+1})$.

15. Except for possibly the case of backtracking in $s_t = s_0$ which does not affect the proof.

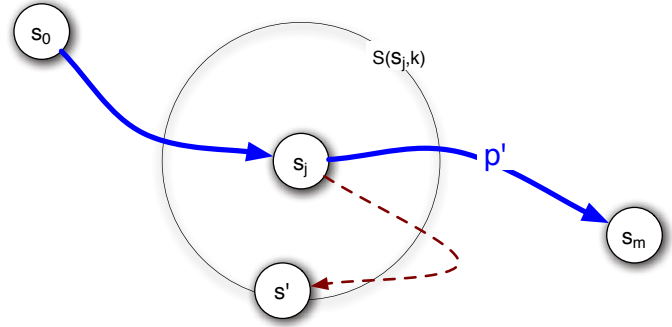

Figure 23: To the proof of Theorem 7.8.

Thus, on the first trial, state $s'$ must have been arrived at via a sequence of forward moves starting in state $s_j$ (shown as the dashed curve in Figure 23). During the forward moves with no $h$-updates, $h(s_{\text{current}}) > h(s_{\text{next}})$ holds. This means that there was a sequence of states traversed by forward moves that started with $s_j$ and ended with $s'$. Since $h_{\text{new}}(s') + k > h(s_j)$, LRTS must have had to backtrack from $s'$ to at least $s_j$. Then it would have updated $h(s_j)$ to $h_{\text{new}}(s_j) \geq h_{\text{new}}(s') + k$ on the first trial. Consequently, state $s'$ cannot be the cause of an update to $h(s_j)$ on the second trial as inequality B.24 will no longer hold, leading to a contradiction. Thus, no updates to $h$ are carried out on the second trial. $\square$

Let us define **solution**$(i)$ as the execution cost of a path $p$ between the start and the goal states such that: (i) $p$ lies fully in the set of states traversed by the agent on trial $i$ and (ii) $p$ minimizes the execution cost.

**Theorem B.1** For any valid $T, \gamma, d$ and an admissible initial heuristic $h_0$, the first trial of LRTS$(d, \gamma, T)$ with systematic or fixed tie-breaking and incremental pruning (see Figure 21) results in **solution**$(1) \leq \frac{h^*(s_0) + T}{\gamma}$.

**Proof.** We will prove this theorem in three steps.

**Notation**. Consider a sequence of states $[s_0, s_1, \ldots, s_n]$ traversed during the first trial. Here state $s_0$ is the start state and state $s_i$ is the current state ($s$) after $i$-th iteration line 27 in LRTS (Figure 20). State $s_n$ is the goal state reached at the end of the first trial. Within iteration $i$ ($0 \leq i \leq n-1$), two states are computed as follows. At the beginning of the iteration (line 6), $s_i$ is the current state $s$. During the forward move (line 23), state $s$ is set to $s_{i+1} = s_{\min}(s_i, d)$.[16] Let us denote the heuristic value of state $s_i$ at the beginning of iteration $i$ as $h_i(s_i)$. After the possible update in line 17, the new heuristic value is $h_{i+1}(s_i)$. Also, let us denote the amount of learning $\ell$ during iteration $i$ as $\ell_i$. More precisely:

$$\ell_i = \begin{cases} h_{i+1}(s_i) - h_i(s_i) & \text{if } h'(s) > h(s) \text{ in line 15,} \\ 0 & \text{otherwise.} \end{cases} \tag{B.25}$$

Under this notation, it is easy to see that the following inequality holds:

$$\gamma \operatorname{dist}(s_i, s_{i+1}) + h_i(s_{i+1}) \leq h_i(s_i) + \ell_i \tag{B.26}$$

---

16. The case of seeing a goal state, i.e., $G \neq \emptyset$, is dealt with in a similar fashion.

when a forward move (line 23) happens. Note that on iteration $i$, state $s_i$ is the only state in which the heuristic value can be updated. Thus, for $\forall s_j \neq s_i \left[ h_{i+1}(s_j) = h_i(s_j) \right]$. In particular[17], $h_{i+1}(s_{i+1}) = h_i(s_{i+1})$ which allows to transform inequality B.26 into:

$$\gamma \operatorname{dist}(s_i, s_{i+1}) + h_{i+1}(s_{i+1}) \leq h_i(s_i) + \ell_i. \tag{B.27}$$

**Forward moves.** We will now consider the case of revisiting states during forward moves. Namely, suppose that in the sequence of states $[s_0, s_1, \ldots, s_n]$, state $s_{i+k} = s_i$ and $k > 0$ is the smallest $k$ such that this equation holds. We will also assume that no backtracking (line 26) took place between states $s_i$ and $s_{i+k}$. The pruning module (lines 23a through 23c in Figure 21) will purge all states between $s_i$ and $s_{i+k}$. Therefore, move $s_{i+k} \to s_{i+k+1}$ will immediately follow move $s_{i-1} \to s_i$ making the new sequence: $s_{i-1} \to s_i (= s_{i+k}) \to s_{i+k+1}$.

For state $s_{i-1}$, the inequality B.27 is:

$$\gamma \operatorname{dist}(s_{i-1}, s_i) + h_i(s_i) \leq h_{i-1}(s_{i-1}) + \ell_{i-1}. \tag{B.28}$$

For state $s_{i+k}$, the inequality B.27 is:

$$\gamma \operatorname{dist}(s_{i+k}, s_{i+k+1}) + h_{i+k+1}(s_{i+k+1}) \leq h_{i+k}(s_{i+k}) + \ell_{i+k}. \tag{B.29}$$

Remembering that $s_i = s_{i+k}$ and the only increase of $h(s_i)$ possible when the algorithm *left* state $s_i$, we conclude that $h_i(s_i) + \ell_i = h_{i+k}(s_{i+k})$. To be consistent with the pruned move sequence $s_{i-1} \to s_{i+k} \to s_{i+k+1}$, we re-write inequality B.28 as:

$$\gamma \operatorname{dist}(s_{i-1}, s_{i+k}) + h_{i+k}(s_{i+k}) - \ell_i \leq h_{i-1}(s_{i-1}) + \ell_{i-1}. \tag{B.30}$$

Adding B.29 and B.30 and re-grouping terms, we arrive at:

$$\gamma \left[ \operatorname{dist}(s_{i-1}, s_{i+k}) + \operatorname{dist}(s_{i+k}, s_{i+k+1}) \right]$$
$$\leq h_{i-1}(s_{i-1}) - h_{i+k+1}(s_{i+k+1}) + \ell_{i-1} + \ell_i + \ell_{i+k}. \tag{B.31}$$

This means that the weighted sum of distances in the sequence of states $[s_{i-1}, s_{i+k}, s_{i+k+1}]$ can be upper bounded by the differences in $h$ in the first and the last states plus the sum of learning amounts for each of the three states. It is easy to show that in the absence of backtracking, inequality B.31 generalizes for the entire *pruned* sequence $[s_0, s_1, \ldots, s_n]$. Note that pruning done to the sequence will manifest itself as gaps in the enumeration of the states. This is similar to the subsequence $s_{i-1}, s_i, s_{i+1}, \ldots, s_{i+k}, s_{i+k+1}$ used earlier in the proof becoming $s_{i-1}, s_{i+k}, s_{i+k+1}$. Let us denote the pruned sequence of indices as $I$. Then inequality B.31 becomes:

$$\gamma \sum_{i \in I} \operatorname{dist}(s_i, s_{i+1}) \leq h_0(s_0) - h_n(s_n) + \sum_{i=0}^{n-1} \ell_i. \tag{B.32}$$

Noticing that $h_0(s_0) \leq h^*(s_0)$ and $h_n(s_n) = 0$ as $s_n$ is a goal state, we arrive at:

$$\gamma \sum_{i \in I} \operatorname{dist}(s_i, s_{i+1}) \leq h^*(s_0) + \sum_{i=0}^{n-1} \ell_i. \tag{B.33}$$

---

17. The special case of $s_i = s_0$ can be dealt with in a simple fashion.

Finally, observing that the total amount of learning is upper bounded by the learning quota $T$ and that the sum of distances along the pruned state sequence is exactly $\mathbf{solution}(1)$, we derive the desired upper bound:

$$\mathbf{solution}(1) \leq \frac{h^*(s_0) + T}{\gamma}. \tag{B.34}$$

**Backtracking.** The finale of the proof lies with showing that backtracking (line 26 in LRTS) does not affect the derivation in the previous section. Suppose, LRTS traversed states $s_i \rightarrow s_{i+1} \rightarrow s_{i+2} \rightarrow \ldots$ and then backtracked to state $s_i$ on iteration $i+k$ (i.e., $s_{i+k} = s_i$). In the pruned solution sequence all states between $s_{i-1}$ and $s_{i+k}$ will be removed from $path$ (line 26) creating a gap in the enumeration: $\ldots, s_{i-1}, s_{i+k}, s_{i+k+1}, \ldots$. Observing that states $s_{i-1}, s_i = s_{i+k}, s_{i+k+1}$ (i) were not backtracked from and (ii) were not among the states removed from the $path$ during backtracking,[18] we conclude that backtracking did not affect the heuristic value of these three states. Therefore, the previous derivation still holds which concludes the proof of the theorem. $\square$

---

18. The fact the three states cannot be among the states removed during backtracking is due to the incremental pruning mechanism in lines 23a, 23b, and 23c of Figure 21. The pruning mechanism clearly separates updates to the heuristic function made during forward and backward moves.